\RequirePackage{fix-cm}
\documentclass[smallextended]{for_arxiv}    

\usepackage{graphicx}
\usepackage{amsmath}
\usepackage{amssymb}
\usepackage{mathtools}
\usepackage{lineno}
\usepackage{color}
\usepackage{url}
\usepackage{hyperref}
\usepackage{subfigure}
\usepackage{multirow}
\usepackage{booktabs}

\usepackage{algorithmic}
\usepackage{array}

\usepackage[final]{review}
\setrevision{2}

\newcommand{\vis}{\ensuremath{V}}
\newcommand{\visval}{\ensuremath{v}}
\newcommand{\visprob}{\ensuremath{\pi_v}}
\newcommand{\visobs}{\ensuremath{\nu}}

\newcommand{\ex}{\ensuremath{e}}
\newcommand{\exset}{\ensuremath{\mathbf{e}}}
\newcommand{\exvect}{\ensuremath{\mathbf{e}}}

\newcommand{\state}{\ensuremath{\mathbf{X}}}
\newcommand{\stateval}{\ensuremath{\mathbf{x}}}
\newcommand{\location}{\ensuremath{\mathbf{L}}}

\newcommand{\vel}{\ensuremath{\mathbf{U}}}

\newcommand{\obsset}{\ensuremath{\mathbf{y}}}
\newcommand{\obsval}{\ensuremath{\mathbf{y}}}
\newcommand{\latent}{\ensuremath{Z}}
\newcommand{\latentset}{\ensuremath{\mathbf{Z}}}

\newcommand{\histval}{\ensuremath{\mathbf{h}}}
\newcommand{\histset}{\ensuremath{\mathbf{h}}}

\newcommand{\obsjval}{\ensuremath{\mathbf{o}}}
\newcommand{\obsjset}{\ensuremath{\mathbf{o}}}

\newcommand{\trans}{\ensuremath{\mathbf{D}}}

\newcommand{\stateCov}{\ensuremath{\mathbf{\Lambda}}}
\newcommand{\mult}{\ensuremath{a}}

\newcommand{\unif}{\ensuremath{u}}
\newcommand{\obsOp}{\ensuremath{\mathbf{P}}}
\newcommand{\obsCov}{\ensuremath{\mathbf{\Sigma}}}

\newcommand{\multq}{\ensuremath{\alpha}}
\newcommand{\smean}{\ensuremath{\boldsymbol{\mu}}}
\newcommand{\scov}{\ensuremath{\mathbf{\Gamma}}}

\newcommand{\q}{\ensuremath{q}}
\newcommand{\param}{\ensuremath{\mathbf{\Theta}}}

\newcommand{\trace}{\textrm{Tr}}

\newcommand{\cost}{\ensuremath{J}}

\newcommand{\idmat}{\ensuremath{\mathbf{I}}}
\newcommand{\zeromat}{\ensuremath{\mathbf{0}}}

\begin{document}

\title{An On-line Variational Bayesian Model for Multi-Person Tracking from Cluttered Scenes\footnote{Support from the EU ERC Advanced Grant VHIA (\#34113) is greatly acknowledged.}}
\author{Sileye Ba \and Xavier Alameda-Pineda \and Alessio Xompero \and Radu Horaud}
\institute{S. Ba \and A. Xompero \and R. Horaud \at INRIA Grenoble Rh\^one-Alpes \\ Montbonnot Saint-Martin, France \and X. Alameda-Pineda \at University of Trento\\ Trento, Italy}
\maketitle

\begin{abstract}
Object tracking is an ubiquitous problem that appears in many applications such as remote sensing, audio processing, computer vision, human-machine 
interfaces, human-robot interaction, etc. Although thoroughly investigated in computer vision, tracking a time-varying number of persons remains a 
challenging open problem. In this paper, we propose an on-line variational Bayesian model for multi-person tracking from cluttered visual observations 
provided by person detectors. The paper has the following contributions. We propose a variational Bayesian framework for tracking an 
unknown and varying number of persons. Our model results in a variational expectation-maximization (VEM) algorithm with closed-form 
expressions both for the posterior distributions of the latent variables and for the estimation of the model parameters. The proposed model exploits 
observations from multiple detectors, and it is therefore multimodal by nature. Finally, we propose to embed both object-birth and object-visibility 
processes in an effort to robustly handle temporal appearances and disappearances. Evaluated on classical multiple person tracking datasets, our method shows competitive 
results with respect to state-of-the-art multiple-object tracking algorithms, such  as the probability hypothesis density (PHD) filter, among others.
\end{abstract}




\section{Introduction}
\label{sec:introduction}
The problem of tracking a varying number of objects is ubiquitous in a number of fields such as remote sensing, computer vision, human-computer 
interaction, human-robot interaction, etc. While several off-line multi-object tracking methods are available, on-line multi-person tracking is still 
extremely challenging \cite{WenhanReview-arXiv}. In this paper we propose an on-line tracking method within the tracking-by-detection (TbD) paradigm, 
which gained popularity in the computer vision community thanks to the development of efficient and robust object detectors~\cite{AndrilukaCVPR2008}. 
Moreover, one advantage of TbD paradigm is the possibility of using linear mappings to link the kinematic (latent) states of the tracked objects to 
the observations issued from the detectors. This is possible because object detectors efficiently capture and filter out the non-linear 
effects, thus delivering detections that are linearly related to the kinematic latent states.

In addition to the difficulties associated to single-object tracking (occlusions, self-occlusions, visual appearance variability, unpredictable 
temporal behavior, etc.), tracking a varying and unknown number of objects makes the problem more challenging because of the following reasons: 
(i)~the observations coming from detectors need to be associated to the objects that generated them, which includes the process of discarding 
detection errors, (ii)~the number of objects is not known in advance and hence it must be estimated, mutual occlusions (not present in 
single-tracking scenarios) must be robustly handled, (iv)~when many objects are present the dimension of the state-space is large, and hence the 
tracker has to handle a large number of hidden-state parameters, (v)~the number of objects varies over time and one has to deal with hidden states of 
varying dimensionality, from zero when there is no visible object, to a large number of detected objects. Note that in this case and if a Bayesian 
setting is being considered, as is often the case, the exact recursive filtering solution is intractable.

In computer vision, previously proposed methodological frameworks for multi-target tracking can be divided into three groups. Firstly, the 
trans-dimensional Markov chain model~\cite{GreenStatisticalScience2003}, where the dimensionality of the hidden state-space is part of the state 
variable. This allows to track a variable number of objects by jointly estimating the number of objects and their kinematic states.  In a computer 
vision scenario, \cite{KhanECCV2004,SmithCVPR2005,YangCVPR2014} exploited this framework for tracking a varying number of objects. The main drawback is that the 
states are inferred  by means of a reversible jump Markov chain Monte Carlo sampling, which is computationally expensive~\cite{GreenBiometrika1995}. 
Secondly, a random finite set multi-target tracking formulation was proposed~\cite{Mahler1998,MahlerIEEESystemMag2004,MahlerIEEESTSP2013}. Initially 
used for radar applications~\cite{Mahler1998}, in this framework the targets are modeled as realizations of a random finite set which is composed of 
an unknown number of elements. Because an exact solution to this model is computationnally intensive, an approximation known as the probability hypothesis density 
(PHD) filter was proposed \cite{MahlerAR2000}. Further sampling-based approximations of random det based filters  were subsequently proposed, 
e.g.~\cite{SidenbladhICIF2003,ClarkTASSP2006,JohansenMCAP2006}. These were exploited in~\cite{MaTSP2006} for tracking a time-varying number of active 
speakers using auditory cues and in~\cite{MaggioTCSVT2008} for multi-target tracking using visual observations. Thirdly, conditional random fields 
(CRF) were also chosen to address multi-target tracking~\cite{YangCVPR2012,Milan-TPAMI-2014,Heili-TIP-2014}. In this case, tracking is casted into an 
energy minimization problem. In another line of research, in radar tracking, other  popular multi-targets tracking model are joint probabilistisc data assocation (JPDA), and multiple hypothesis filters \cite{BarShalom2009}.

In this paper we propose an on-line variational Bayesian framework for tracking an unknown and varying number of persons. Although variational model 
are very popular in machine learning, their use in computer vision for object tracking has been limited to tracking situation involving a fixed number 
of targets~\cite{VermaakCVPR2003}. Variational Bayes methods approximate the joint a posteriori distribution of the latent variables by a separable 
distribution~\cite{SmidlVariationalBayesSpringer2006,bishop2007}. In an on-line tracking scenario, where only causal (past) observations can be used, 
this translates into approximating the filtering distribution. This is in strong contrast with off-line trackers that use both past and future 
observations. The proposed tracking algorithm is therefore modeling the a posteriori distribution of the hidden states given all past observations. 
Importantly, the proposed framework leads to closed-form expressions for the posterior distributions of the hidden variables and for the model 
parameters, thus yielding an intrinsically efficient filtering procedure implemented via an variational EM (VEM) algorithm.  In addition, a 
\textit{clutter target} is defined so that spurious observations, namely detector failures, are associated to this target and do not contaminate the 
filtering process. Furthermore, our formalism allows to integrate in a principled way heterogeneous observations coming from various detectors, e.g, 
face, upper-body, silhouette, etc. Remarkably, objects that come in and out of the field of view, namely object appearance and disappearance, are 
handled by object birth and visibility processes. In details, we replace the classical death process by a visibility process which allows to put 
asleep tracks associated with persons that are no longer visible. The main advantage is that these tracks can be awaken as soon as new observations 
match their appearance with high confidence. Summarizing, the paper contributions are:
\begin{itemize}
\item Cast the problem of tracking a time-varying number of people into a variational Bayes formulation, which approximates the a posteriori 
filtering distribution by a separable distribution;
\item A VEM algorithm with closed-form expressions, thus inherently efficient, for the update of the a posteriori distributions and the 
estimation of the model parameters from the observations coming from different detectors;
\item An object-birth and an object-visibility process allowing to handle person appearance and disappearance due either to occlusions or people 
leaving the visual scene;
\item A thorough evaluation of the proposed method compared with the state-of-the-art in two datasets, the cocktail party dataset and a dataset 
containing several sequences traditionally used in the computer vision community to evaluate multi-person trackers.
\end{itemize}

The remainder of this paper is organized as follows. Section \ref{sec:related-work} reviews previous work relevant to our work method. Section 
\ref{sec:mot-model} details the proposed Bayesian model and a variational model solution is presented in Section~\ref{sec:variational-approximation}. 
In Section~\ref{sec:birth-visibility-process}, we depict the birth and visibility processes allowing to handle an unknown and varying number of 
persons. Section \ref{sec:experiments} describes results of experiments and benchmarks to assess the quality of the proposed method. Finally, Section 
\ref{sec:conclusions} draws some conclusions.
%
%
%
\section{Related Work}
\label{sec:related-work}
Generally speaking, object tracking is the temporal estimation of the object's kinematic state. In the context of image-based tracking, the object state 
is typically a parametrization of its localization in the (2D) image plane. In computer vision, object tracking has been thoroughly investigated 
\cite{Yilmaz2006}. 
Objects of interest could be people, faces, hands, vehicles, etc. According to the considered number of objects to be 
tracked, tracking can be classified into single-object tracking, fixed-number multi-object tracking, and varying-number multi-object tracking. 

Methods for single object tracking consider only one object and usually involve an initialization step, a state update step, and a 
reinitialization step allowing to recover from failures. Practical initialization steps are based on generic object detectors allowing to scan the 
input image in order to find the object of interest~\cite{FelzenszwalbTPAMI2010,ZhuRamananCVPR2012}. Object detectors can be used for the 
 reinitialization step as well. However, using generic object detectors is problematic when other objects of the same kind than the tracked object 
are present in the visual scene. In order to resolve such ambiguities, different complementary appearance models have been proposed, such as object 
templates, color appearance models, edges (image gradients) and texture, (e.g. Gabor features and histogram of gradient orientations). Regarding 
the update step, the current state can be estimated from previous states and observations with either deterministic~\cite{ComaniciuPAMI2002} or
probabilistic~\cite{ArulampalamIEEE-SP2002} methods.

Even if it is still a challenging problem, tracking a single object is very limited in scope. Rapidly, the computer vision community drove its 
attention towards fixed-number multi-object tracking \cite{SarkaICIF2004}. Additional difficulties are encountered when tracking multiple objects. 
Firstly, there is an increase of the tracking state dimensionality as the multi-object tracking state dimensionality is the single object state 
dimensionality multiplied by the number of tracked objects. Secondly, associations between observations and objects are required. Since the 
observation-to-object association problem is combinatorial~\cite{YanCVPR2006,BarShalom2009}, it must be carefully addressed when the number of objects 
and of observations are large. Thirdly, because of the presence of multiple targets, tracking methods have to be robust also to mutual occlusions.

In most practical applications, the number of objects to be tracked, is not only unknown, but it also varies over time. Importantly, tracking 
a time-varying number or objects requires an efficient mechanism to add new objects entering the field of view, and to remove objects that moved 
away. In a probabilistic setting, these mechanisms are based on birth and death processes. Efficient multi-object algorithms have to be developed 
within principled methodologies allowing to handle hidden states of varying dimensionality. In computer vision, the most popular methods are based on 
conditional random fields~\cite{BaeYoonCVPR2014,Milan-TPAMI-2014,Heili-TIP-2014,Pirsiavash-CVPR-2016}, on random finite 
sets~\cite{MahlerIEEESTSP2013,MaTSP2006,MaggioTCSVT2008} or on the trans-dimensional Markov 
chain~\cite{GreenStatisticalScience2003,KhanECCV2004,SmithCVPR2005,YangCVPR2014}. 
\cite{YangCVPR2014} presents an interesting approach where occlusion state of a tracked person is explicitly modeled 
in the tracked state and used for observation likelihood computation. 
Less popular but successful methodologies include the Bayesian multiple blob 
tracker of \cite{IsardICCV2001}, the boosted particle filter for multi-target tracking of \cite{OkumaECCV2004} and the Rao-Blackwellized filter for 
multiple objects tracking \cite{SarkaElsevierIF2005}, graph based representation for multi-object tracking \cite{Zamir-CVPR-2012,Wen-CVPR-2014}. It has to be noticed in other communities, such as radar tracking, multi-object tracking has 
been deeply investigated. Many models have been proposed such as the probabilistic data association filter (PDAF), the joint PDAF,  multiple 
hypothesis tracking \cite{BarShalom2009}. However, the differences between multi-object tracking in radar and in computer vision are 
mainly two. On the one hand, most tracking method for radar consider point-wise objects, modeling a punctual latent state, whereas in computer vision 
objects are represented using bounding boxes in addition to the punctual coordinates. On the other hand, computer vision applications benefit 
from the use of visual appearance, which is mainly used for object identification~\cite{PerezECCV2002}.

Currently available multi-object tracking methods used in computer vision applicative scenarios suffer from different drawbacks. CRF-based approaches 
are naturally non-causal, that is, they use both past and future information. Therefore, even if they have shown high robustness to clutter, they are 
only suitable for off-line applications when smoothing (as opposite to filtering) techniques can be used. \addnote[phd]{1}{PHD filtering techniques report good 
computational efficiency, but they are inherently limited since they provide non-associated tracks. In other words, these techniques require an 
external method in order to associate observations and tracks to objects}. Finally, even if trans-dimensional MCMC based tracking techniques are able to associate tracks to objects 
using only causal information, they are extremely complex from a computational point of view, and their performance is very sensitive to the sampling 
procedure used. In contrast, the variational Bayesian framework we propose associates tracks to previously seen objects and creates new tracks in an 
unified framework that filters past observations in an intrinsically efficient way, since all the steps of the algorithm are expressed in closed-form. 
Hence the proposed method robustly and efficiently tracks a varying and unknown number of persons from a combination of image detectors.

\section{Variational Bayesian Multiple-Person Tracking}
\label{sec:mot-model}
\subsection{Notations and Definitions}
\label{subsec:variables-definition}
We start by introducing our notations. Vectors and matrices are in bold $\mathbf{A}$, $\mathbf{a}$, scalars are in italic $A$, $a$.
\addnote[random-var-not]{1}{In general random variables are denoted with upper-case letters, e.g. $\mathbf{A}$ and $A$, and 
their realizations with lower-case letters, e.g. $\mathbf{a}$ and $a$.} 

Since the objective is to track multiple persons whose number may vary over time, we assume that there is a maximum number of people, denoted by $N$, 
that may enter the visual scene. This parameter is necessary in order to cast the problem at hand into a finite-dimensional state 
space, consequently $N$ can be arbitrarily large. A track $n$ at time $t$ is associated to the \textit{existence} binary variable $\ex_{tn}$ taking 
the value $\ex_{tn}=1$ if the person has already been seen and $\ex_{tn}=0$ otherwise. The vectorization of the existence variables at time $t$ is 
denoted by $\exset_t=(\ex_{t1},...,\ex_{tN})$ and their sum, namely the effective number of tracked persons at $t$,  is denoted by $N_t = \sum_{n=1}^N 
\ex_{tn}$. The existence variables are assumed to be observed in sections~3 and~4; Their inference, grounded in a track-birth stochastic process, is discussed in section 5.

The kinematic state of person $n$ is a random vector $\state_{tn}=(\location_{tn}^\top,\vel_{tn}^\top)^\top\in\mathbb{R}^6$, where 
$\location_{tn}\in\mathbb{R}^4$ 
is the person location, i.e., 2D image position, width and height, and  $\vel_{tn}\in\mathbb{R}^2$ is the person velocity in the image plane. The 
multi-person state random vector is denoted by $\state_t=(\state_{t1}^\top,\ldots,\state_{tN}^\top)^\top\in\mathbb{R}^{6N}$. Importantly, the 
kinematic state is described by a set of hidden variables which must be robustly estimated.

\begin{figure}[t]
\begin{center}
\subfigure[]{\label{subfig:observations1}\includegraphics[width=0.4\textwidth]{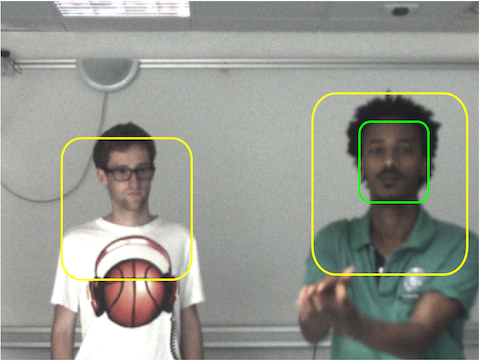} }
\subfigure[]{\label{subfig:observations2}\includegraphics[width=0.4\textwidth]{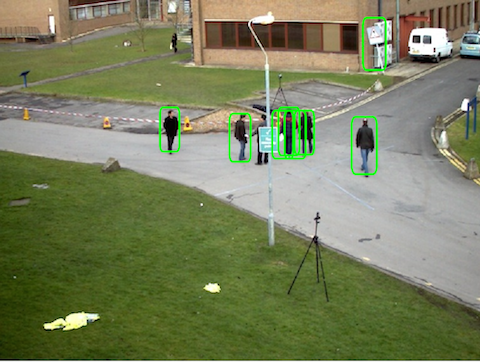}  }
\caption{Examples of detections used as observations by the proposed person tracker: upper-body, face (a), and full-body (b) detections. Notice that 
one of the faces was not detected and that there is a false full-body detection in the background.}
\label{fig:observations}
\end{center}
\end{figure}

In order to ease the challenging task of tracking multiple persons with a single static camera, we assume the existence of $I$ detectors, each of 
them providing $K_t^i$ localization observations at each time $t$, with $i\in[1 \dots I]$. Fig.~\ref{fig:observations} provides examples of face and 
upper-body detections (see Fig.~\ref{subfig:observations1}) and of full-body detections (see Fig.~\ref{subfig:observations2}). The $k$-th 
localization observation gathered by the $i$-th detector at time $t$ is denoted by $\obsval_{tk}^i\in\mathbb{R}^4$, and represents the location (2D 
position, width, height) of a person in the image. The set of observations provided by detector $i$ at time $t$ is denoted by 
$\obsval_t^i=\{\obsval_{tk}^i\}_{k=1}^{K_t^i}$, and the observations provided by all the detectors at time $t$ is 
denoted by $\obsset_t = \{\obsset_t^i\}_{i=1}^I$.  Associated to each localization detection $\obsval_{tk}^i$, there is a photometric description of 
the person's appearance, denoted by $\histval_{tk}^i$. This photometric observation is extracted from the bounding box of $\obsval_{tk}^i$. 
Altogether, the localization and photometric observations constitute  the raw observations $\obsjval_{tk}^i=(\obsval_{tk}^{i},\histval_{tk}^{i})$ used 
by our tracker. \addnote[definitions]{1}{Analogous 
definitions to $\obsval_t^i$ and $\obsval_t$ hold for $\histset_t^i=\{\histval_{tk}^i\}_{k=1}^{K_t}$, $\histset_t=\{\histset_t^i\}_{i=1}^I$, 
$\obsjset_t^i=\{\obsjval_{tk}^i\}_{k=1}^{K_t}$ and $\obsjset_t=\{\obsjset_t^i\}_{i=1}^I$.} \addnote[prob-sets]{1}{Importantly, when we write the 
probability of a set of random variables, we refer to the joint probabilities of all random variables in that set. For instance: $p(\obsjset_t^i) = 
p(\obsjval_{t1}^i,\ldots,\obsjval_{tK_t^i}^i)$.}

We also define an observation-to-person assignment (hidden) variable $\latent_{tk}^i$ associated with each observation $\obsjval_{tk}^i$. Formally, 
$\latent_{tk}^i$ is a categorical variable taking values in the set $\{1 \ldots N\}$: $Z_{tk}^i=n$ means that $\obsjval_{tk}^i$ is associated to 
person $n$. $\latentset_t^i$ and $\latentset_t$ are defined in an analogous way to $\obsset_t^i$ and $\obsset_t$. These assignment variables 
can be easily used to handle false detections. Indeed, it is common that a detection corresponds to some clutter instead of a person. We cope with 
these false detections by defining a \textit{clutter} target. In practice, the index $n=0$ is assigned to this clutter target, which is always 
visible, i.e. $e_{t0}=1$ for all $t$. Hence, the set of possible values for $Z_{tk}^i$ is extended to $\{0\}\cup\{1 \ldots N\}$, and 
$Z_{tk}^i=0$ means that observation $\obsjval_{tk}^i$ has been generated by clutter and not by a person. The practical consequence of adding a 
clutter track is that the observations assigned to it play no role in the estimation of the parameters of the other tracks, thus leading to 
estimation rules inherently robust to outliers.

\subsection{The Proposed Bayesian Multi-Person Tracking Model}
\begin{figure}[t]
 \centering
 \includegraphics[height=5cm]{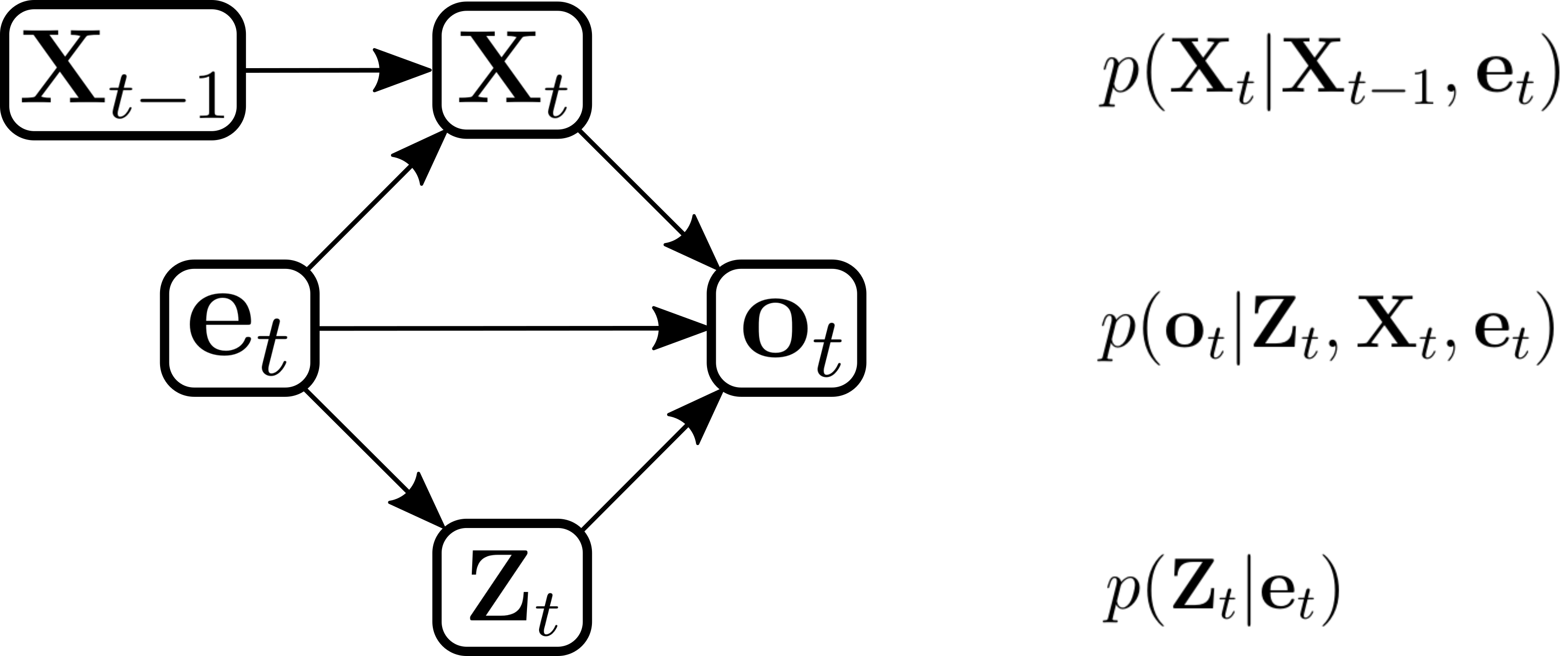}
 \caption{Graphical representation of the proposed multi-target tracking probabilistic model.\label{fig:graph_model}}
\end{figure}

The on-line multi-person tracking problem is cast into the estimation of the filtering distribution of the hidden variables given the 
causal observations $p(\latentset_t,\state_t|\obsjset_{1:t}, \exset_{1:t})$, where $\obsjset_{1:t}= \{\obsjset_1, \dots, \obsjset_t \}$. The filtering 
distribution can be rewritten as:
\begin{equation}
\label{eq:posterior-gen}
p(\latentset_t,\state_t|\obsjset_{1:t}, \exset_{1:t}) = \frac{p(\obsjset_t|\latentset_t,\state_t,\obsjset_{1:t-1},\exset_{1:t}) 
p(\latentset_t,\state_t|\obsjset_{1:t-1}, \exset_{1:t}) }{p(\obsjset_t|\obsjset_{1:t-1}, \exset_{1:t})}.
\end{equation} 
Importantly, we assume that the observations at time $t$ only depend on the hidden and visibility variables at time $t$. 
Therefore~\eqref{eq:posterior-gen} writes:
\begin{equation}
p(\latentset_t,\state_t|\obsjset_{1:t}, \exset_{1:t}) =  
\frac{p(\obsjset_t|\latentset_t,\state_t,\exset_t)p(\latentset_t|\exset_t)p(\state_t|\obsjset_{1:t-1}, 
\exset_{1:t}) }{p(\obsjset_t|\obsjset_{1:t-1}, \exset_{1:t})}\label{eq:posterior}.
\end{equation}
The denominator of \eqref{eq:posterior} only involves observed variables and therefore its evaluation is not necessary as long as one can normalize 
the expression arising from the numerator. Hence we focus on the three terms of the latter, namely the 
observation model $p(\obsjset_t|\latentset_t,\state_t,\exset_t)$, the observation-to-person assignment prior distribution 
$p(\latentset_t|\exset_t)$ and the dynamics of the latent state $p(\state_t|\state_{t-1},\exset_{t})$, which appear when marginalizing the predictive distribution $p(\state_t|\obsjset_{1:t-1}, 
\exset_{1:t})$ with respect to $\state_{t-1}$. 
Figure~\ref{fig:graph_model} shows a graphical schematic representation of the proposed probabilistic model.

\subsubsection{The Observation Model}
\label{subsubsec:observation-likelihood}
The joint observations are assumed to be independent and identically distributed:
\begin{equation}
p(\obsjset_t|\latentset_t,\state_t,\exset_t) = \prod_{i=1}^I\prod_{k=1}^{K_t^i} p(\obsjval_{tk}^i|\latent_{tk}^i,\state_t,\exset_t).
\label{eq:likelihoods}
\end{equation}
In addition, we make the reasonable assumption that, while localization observations depend both on the assignment variable and kinematic state, the 
appearance observations only depend on the assignment variable, that is the person identity, but not on his/her kinematic state. We also assume 
the localization and appearance observations to be independent given the hidden variables. Consequently, the observation likelihood of a single joint 
observation can be factorized as:
\begin{eqnarray}
p(\obsjval_{tk}^i|\latent_{tk}^i,\state_t,\exset_t) &=& p(\obsval_{tk}^i,\histval_{tk}^i|\latent_{tk}^i,\state_t,\exset_t)\\ \nonumber 
&=&p(\obsval_{tk}^i|\latent_{tk}^i,\state_t,\exset_t)p(\histval_{tk}^i|\latent_{tk}^i,\exset_t).
\end{eqnarray}
The localization observation model is defined depending on whether the observation is generated by clutter or by a person:
\begin{itemize}
\item If the observation is generated from clutter, namely $\latent_{tk}^i =0$, the variable $\obsval_{tk}^i$ follows an uniform distribution with 
probability density function $\unif(\obsval_{tk}^i)$;
\item If the observation is generated by person $n$, namely $\latent_{tk}^i =n$, the variable $\obsval_{tk}^i$ follows a Gaussian distribution 
with mean $\obsOp^i \state_{tn}$ and covariance $\obsCov^i$: $\obsval_{tk}^i\sim g(\obsval_{tk}^i;\obsOp^i \state_{tn},\obsCov^i)$
\end{itemize}
The linear operator $\obsOp^i$  maps the kinematic state vectors onto the $i$-th space of observations. For example, when $\state_{tn}$ represents 
the upper-body kinematic state (upper-body localization and velocity) and $\obsval_{tk}^i$ represents the upper-body localization observation, 
$\obsOp^i$ is a projection which, when applied to a state vector, only retains the localization components of the state vector. When $\obsval_{tk}^i$ 
is a face localization observation, the operator $\obsOp^i$ is a composition of the previous projection, and an affine transformation mapping an 
upper-body bounding-box onto its corresponding face bounding-box. Finally, the full observation model is compactly defined by
\begin{equation} \label{eq:motion-likelihood}
\!\! p(\obsval_{tk}^i|\latent_{tk}^i =n,\state_{t},\exset_t)=  \unif(\obsval_{tk}^i)^{1-\ex_{tn}} \left(\unif(\obsval_{tk}^i)^{\delta_{0n}}  
g(\obsval_{tk}^i; \; \obsOp^i \state_{tn},\obsCov^i) ^{1-\delta_{0n}}\right)^{\ex_{tn}},
\end{equation}
where $\delta_{ij}$ stands for the Kronecker function.

The appearance observation model is also defined depending on whether the observations is clutter or not. When the observation is generated by 
clutter, the appearance observation follows a uniform distribution with density function $\unif(\histval_{tk}^i)$. When the observation is generated 
by person $n$, the appearance observation follows a Bhattacharya distribution with density defined as 
$$b(\histval_{tk}^i;\histval_n)=\frac{1}{W_\lambda} \exp(-\lambda d_B(\histval_{tk}^i,\histval_n) ),$$
where $\lambda$ is a positive skewness parameter, $d_B(.,.)$ is the Battacharya distance between histograms, $\histval_n$ is the $n$-th person's 
reference appearance model\footnote{It should be noted that the normalization constant  $W_\lambda = \int_{\sum_k \histval_k=1} \exp(-\lambda 
d_B(\histval,\histval_n) ) \mbox{d}\histval$ can be exactly computed only for histograms with dimension lower than 3. In practice $W_\lambda$ is 
approximated using Monte Carlo integration.}. This gives the following
compact appearance observation model:
\begin{eqnarray}\label{eqn:color-likelihood}
 p(\histval_{tk}^i|\latent_{tk}^i=n,\state_t,\exset_t) = \unif(\histval_{tk}^i)^{1-\ex_{tn}} (\unif(\histval_{tk}^i)^{\delta_{0n}} 
b(\histval_{tk}^i;\histval_n)^{1-\delta_{0n}} )^{\ex_{tn}}.
\end{eqnarray}

\subsubsection{The Observation-to-Person Prior Distribution}
\label{subsubsec:assignment-distribution}
The joint distribution of the assignment variables factorizes as:
\begin{equation}
p(\latentset_t|\exset_t)= \prod_{i=1}^I\prod_{k=1}^{K_t^i} p(\latent_{tk}|\exset_t).
\label{eq:Zset-prior}
\end{equation}
When observations are not yet available, given existence variables $\exset_t$, the assignment variables $\latent_{tk}^i$ are assumed to follow multinomial distributions defined as:
\begin{equation}
\label{eqn:assignment-distribution}
p(\latent_{tk}^i =n|\exvect_t) =  \ex_{tn}a_{tn}^i \quad\text{with}\quad\sum_{n=0}^Ne_{tn}a_{tn}^i = 1.
\end{equation}
Because $\ex_{tn}$ takes the value $1$ only for actual persons, the probability to assign an observation to a non-existing person is null.

\subsubsection{The Predictive Distribution}
\label{subsubsec:dynamics}
The kinematic state predictive distribution represents the probability distribution of the kinematic state at time $t$ given the observations up to time  
$t-1$ and the existence variables $p(\state_t|\obsjset_{1:t-1}, \exset_{1:t})$. The predictive distribution is mainly driven by the dynamics of people's kinematic states, which are modeled consdering two hypothesis. Firstly the kinematic state dynamics follow a first-order Markov chain, meaning that 
the state $\state_t$ only depends on state $\state_{t-1}$. Secondly, the person locations do not influence each other's dynamics, meaning that there 
is one first-order Markov chain for each person. Formally, this can be written as:
\begin{equation}
 p(\state_{t} | \state_{1:t-1}, \exset_{1:t}) = p(\state_t | \state_{t-1}, \exset_t) = \prod_{n=1}^N p(\state_{tn} | \state_{t-1n}, 
\ex_{tn}).\label{eq:prior-X}
\end{equation}
The immediate consequence is that the posterior distribution computes:
\begin{equation}\label{eq:dynamics}
\! p(\state_t|\obsjset_{1:t-1}, \exset_{1:t})=   \int \left(\prod_{n=1}^ N  p(\state_{tn}|\stateval_{t-1n},\ex_{tn}) \right) p(\stateval_{t-1}|\obsjset_{1:t-1}, 
\exset_{1:t-1}) d\stateval_{t-1}.\!
\end{equation}
For the model to be complete, $p(\state_{tn} | \state_{t-1,n}, \ex_{tn})$ needs to be defined. The temporal evolution 
of the kinematic state $\state_{tn}$ is defined as:
\begin{equation}
p(\state_{tn} = \stateval_{tn} |\state_{t-1,n} =\stateval_{t-1,n},\ex_{tn}) = \unif(\stateval_{tn})^{1-\ex_{tn}} g(\stateval_{tn}; \; \trans 
\stateval_{t-1,n},\stateCov_n) ^{\ex_{tn}},
\label{eq:dynamical-model}
\end{equation}
where $\unif(\stateval_{tn})$ is a uniform distribution over the motion state space, $g$ is a Gaussian probability density function, 
$\trans$ represents the dynamics transition operator, and $\stateCov_n$ is a covariance matrix accounting for uncertainties on the state dynamics. 
The 
transition operator 
is
defined as:
$$
\trans = \left (
\begin{array}{cc}
  \idmat_{4\times 4}  & 
      \begin{array}{c}
      \idmat_{2\times 2}  \\
      \zeromat_{2\times 2} \\
      \end{array}
  \\
  \zeromat_{2\times4} & \idmat_{2\times 2} \\
 \end{array}
 \right) =
 \left(
\begin{array}{cccccc}
1 & 0 & 0 & 0 & 1 & 0 \\
0 & 1 & 0 & 0 & 0 & 1 \\
0 & 0 & 1 & 0 & 0 & 0 \\
0 & 0 & 0 & 1 & 0 & 0 \\
0 & 0 & 0 & 0 & 1 & 0 \\
0 & 0 & 0 & 0 & 0 & 1 
\end{array}
 \right).
$$
In other words, the dynamics of an existing person $n$ is \textit{either} follows a Gaussian with mean vector $\trans \state_{t-1,n}$ and covariance 
matrix $\stateCov_n$, 
\textit{or} a uniform distribution if person $n$ does not exist. 
The complete set of parameters of the proposed model is denoted with 
$\param=\big( \{\obsCov^i\}_{i=1}^I,\;\{\stateCov_n\}_{n=1}^N,\mathbf{A}_{1:t}\big )$, with $\mathbf{A}_{t}=\{a_{tn}^i\}_{n=0,i=1}^{N,I}$.

\section{Variational Bayesian Inference}
\label{sec:variational-approximation}
Because of the combinatorial nature of the observation-to-person assignment problem, a direct optimization of the filtering distribution 
\eqref{eq:posterior}
with respect to the hidden variables is intractable. \addnote[variational-approx]{1}{We propose to overcome this problem via a variational 
Bayesian inference method. The principle of this family of methods is to approximate  the intractable filtering distribution 
$p(\latentset_t,\state_t|\obsjset_{1:t},\exset_{1:t})$ by a separable distribution, e.g. $q(\latentset_t)\prod_{n=0}^Nq(\state_{tn})$. According to 
the variational Bayesian formulation \cite{SmidlVariationalBayesSpringer2006,bishop2007}, given the observations and the parameters at the previous 
iteration $\param^\circ$, the optimal approximation has the following general expression:
\begin{align}
 \label{eqn:variational-expectations-assign-generic}
 \log q(\latentset_{t}) &=\mathbf{E}_{q(\state_t)}\left\{\log 
p(\latentset_t,\state_t|\obsjset_{1:t},\exvect_{1:t},\param^\circ)\right\},\\
\label{eqn:variational-expectations-motion}
 \log q(\state_{tn}) &= \mathbf{E}_{q(\latentset_t)\prod_{m\neq n}q(\state_{tm})}\left\{\log p(\latentset_t,\state_t|\obsjset_{1:t}, 
\exvect_{1:t},\param^\circ)\right\}. 
\end{align}
In our particular case, when these two equations are put together with the probabilistic model defined 
in~(\ref{eq:likelihoods}),~(\ref{eq:Zset-prior}) and~(\ref{eq:prior-X}), the expression of $q(\latentset_t)$ factorizes further into:
\begin{align}
 \label{eqn:variational-expectations-assign}
 \log q(\latent_{tk}^i) &=\mathbf{E}_{q(\state_t)}\left\{\log 
p(\latent_{tk}^i,\state_t|\obsjset_{1:t},\exvect_{1:t},\param^\circ)\right\},
\end{align}
Note that this equation leads to a finer factorization that the one we imposed. This behavior is typical of variational Bayes methods 
in which a very mild separability assumption can lead to a much finer factorization when combined with priors over hidden states and latent variables, i.e.~(\ref{eq:likelihoods}),~(\ref{eq:Zset-prior}) and~(\ref{eq:prior-X}). The final factorization writes:
\begin{equation}
p(\latentset_t,\state_t|\obsjset_{1:t},\exset_{1:t})\approx \prod_{i=1}^I \prod_{k=0}^{K_t^i} q(\latent_{tk}^i) 
\prod_{n=0}^{N} q(\state_{tn}).\label{eq:variational-approximation}
\end{equation}}

Once the posterior distribution over the hidden variables is computed (see below), the optimal parameters are estimated using $\hat{\param} =\arg\max 
_{\param} \cost(\param,\param^\circ)$ with $\cost$ defined as:
\begin{equation}
\cost(\param,\param^\circ)=\mathbf{E}_{q(\latentset_t,\state_t)}\left\{\log p(\latentset_t,\state_t,\obsjset_{1:t}| 
\exvect_{1:t},\param,\param^\circ)\right\}.
\label{eq:variational-maximization}
\end{equation}

To summarize, the proposed solution for multi-person tracking is an on-line variational EM algorithm. Indeed, the 
factorization \eqref{eq:variational-approximation} leads to a variational EM in which the E-step consists of 
computing~\eqref{eqn:variational-expectations-assign} and~\eqref{eqn:variational-expectations-motion} and the M-step consists of maximizing the 
expected complete-data log-likelihood \eqref{eq:variational-maximization} \addnote[param-est]{1}{with respect to the parameters.} However, as is
detailed below, for stability reasons the covariance matrices are not estimated with the variational inference framework, but set to a fixed value. The expectation and 
maximization steps of the algorithm are now detailed.

\subsection{E-Z-Step}
\label{subsec:E-step-assigment}
The estimation of $q(\latent_{tk}^i)$ is carried out by developing the expectation (\ref{eqn:variational-expectations-assign}). More derivation details can be found in~\ref{subsubsec:E-steps-assignment-app}, which yields the following formula:
\begin{equation}
q(\latent_{tk}^i=n) =  \multq_{tkn}^i,
\end{equation}
where 
\begin{equation}
 \multq_{tkn}^i = \frac{\ex_{tn} \epsilon_{tkn}^i a_{tn}^i}{\sum_{m=0}^N \ex_{tm}  \epsilon_{tkm}^i a_{tn}^i},
\end{equation}
and $\epsilon_{tkn}^i$ is defined as:
\begin{equation}
\epsilon_{tkn}^i = \left\{\begin{array}{ll} \unif(\obsval_{tk}^i) \unif(\histval_{tk}^i) & n=0, \\
 g(\obsval_{tk}^i,\obsOp^i \smean_{tn},\obsCov^i) e^{-\frac{1}{2} \trace\left(\obsOp^{i\top} \left(\obsCov^i\right)^{-1} \obsOp^i 
\scov_{tn}\right)} b(\histval_{tk}^i;\histval_n) & n \neq 0,
\end{array}\right.
\end{equation}
where $\trace (\cdot)$ is the trace operator and $\smean_{tn}$ and $\scov_{tn}$ are defined by
\eqref{eqn:q-motion-parameters-mean} and~\eqref{eqn:q-motion-parameters-cov} below. Intuitively, this approximation shows that the 
assignment of an observation to a person is based on spatial proximity between the observation localization and the person localization, and the 
similarity between the observation's appearance and the person's reference appearance.
 
\subsection{E-X-Step}
\label{subsec:E-step-state}
The estimation of $q(\state_{tn})$ is derived from~(\ref{eqn:variational-expectations-motion}). Similarly to the previous posterior
distribution, the mathematical derivations are provided in~\ref{subsubsec:E-step-motion-app}, and boil down to the following formula:
\begin{equation}
q(\state_{tn}) = \unif(\state_{tn})^{1-\ex_{tn}} g(\state_{tn};\smean_{tn},\scov_{tn} )^{\ex_{tn}}, \label{eqn:q_x}
\end{equation}
where the mean vector $\smean_{tn}$ and the covariance matrix $\scov_{tn}$ are given by
\begin{align} 
\label{eqn:q-motion-parameters-mean}
\scov_{tn}  &= \Big( \sum_{i=1}^I \sum_{k=0}^{K_t^i} \multq_{tkn}^i \left(\obsOp^{i\top} \left(\obsCov^{i}\right)^{-1} \obsOp^i\Big)  + (\trans 
\scov_{t-1,n} \trans^\top + \stateCov_n)^{-1}  \right)^{-1} \\
\label{eqn:q-motion-parameters-cov}
\smean_{tn} &= \scov_{tn}\Big( \sum_{i=1}^I \sum_{k=0}^{K_t^i} \multq_{tkn}^i \obsOp^{i\top} \left(\obsCov^{i}\right)^{-1} \obsval_{tk}^i + (\trans 
\scov_{t-1,n} \trans^\top + \stateCov_n)^{-1}\trans \smean_{t-1,n} \Big).
\end{align}
We note that the variational approximation of the kinematic-state distribution reminds the Kalman filter solution of a linear dynamical system with 
mainly one difference: in our solution~\eqref{eqn:q-motion-parameters-mean} and \eqref{eqn:q-motion-parameters-cov}, the mean vectors and covariance 
matrices are computed with the observations weighted by $\multq_{tkn}^i$ (see \eqref{eqn:q-motion-parameters-mean} and \eqref{eqn:q-motion-parameters-cov}).
%

\subsection{M-step}
\label{subsec:m-steps}
Once the posterior distribution of the hidden variables is estimated, the optimal parameter values can be estimated via 
maximization of $\cost$ defined in~\eqref{eq:variational-maximization}. The M-step allows 
to estimate the model parameter.

Regarding the parameters of the a priori observation-to-object assignment $\mathbf{A}_t$ we compute:
\begin{equation}
 J(a_{tn}^i) = \sum_{k=1}^{K_t^i} e_{tn}\alpha_{tkn}^i\log(e_{tn}a_{tn}^i) \quad \text{s.t.} \quad \sum_{n=0}^N e_{tn}a_{tn}^i = 1,
\end{equation}
and trivially obtain:
\begin{equation}
 a_{tn}^i = \frac{e_{tn}\sum_{k=1}^{K_t^i}\alpha_{tkn}^i}{ \sum_{m=0}^N e_{tm}\sum_{k=1}^{K_t^i}\alpha_{tkm}^i}.
\end{equation}

The M-Step for observation covariances corresponds to the estimation of $\obsCov^i$. This is done by maximizing 
$$\cost(\obsCov^i)= \sum_{k=1}^{K_t^i} \sum_{n=1}^N \ex_{tn} \multq_{tkn}^i \log(\obsval_{tk}^i,\obsOp^i \state_{tn},\obsCov^i) $$
with respect to $\obsCov^i$. Differentiating $\cost(\obsCov^i)$ with respect to $\obsCov^i$ and equating to zero gives:
\begin{equation}
\obsCov^i=\frac{1}{K_t^i N} \sum_{k=1}^{K_t^i} \sum_{n=1}^N
\ex_{tn} \multq_{tkn}^i \left( \obsOp^i \scov_{tn} {\obsOp^i}^\top + (\obsval_{tk}^i -\obsOp^i \smean_{tn})(\obsval_{tk}^i -\obsOp^i 
\smean_{tn})^\top  \right)
\end{equation}

The M-Step for kinematic state dynamics covariances corresponds to the estimation of $\stateCov_n$ for a fixed $n$. This done by maximizing cost 
function 
$$\cost(\stateCov_n)=\mathbf{E}_{\q(\state_{tn}|\ex_{tn})}[\log g(\state_{tn};\trans \smean_{t-1n},\trans \scov_{tn} \trans^\top + \stateCov_n)^{\ex_{tn}})].$$
Equating differential of the cost $\cost(\stateCov_n)$ to zeros gives:
\begin{equation}
 \stateCov_n =   \trans \scov_{t-1n} \trans ^\top + \scov_{tn} 
 +  (\smean_{tn}-\trans \smean_{t-1,n})(\smean_{tn}-\trans \smean_{t-1,n})^\top 
\end{equation}

It is worth noticing that, in the current filtering formalism, the formulas for $\obsCov^i$ and $\stateCov_n$ are instantaneous, i.e., they are 
estimated only from the observations at time $t$. The information at time $t$ is usually insufficient to obtain stable values for these matrices. 
Even if estimating $\obsCov^i$ and $\stateCov_n$ is suitable in a parameter learning scenario where the tracks are provided, we noticed that in 
practical tracking scenarios, where the tracks are unknown, this does not yield stable results. Suitable priors on the temporal dynamics of the 
covariance parameters are required. Therefore, in this paper we assume that the observation and dynamical model covariance matrices are fixed.

\section{Person-Birth and Person-Visibility Processes}
\label{sec:birth-visibility-process}
Tracking a time-varying number of targets requires procedures to create tracks when new targets enter the scene and to delete tracks when 
corresponding targets leave the visual scene. In this paper, we propose a statistical-test based birth process that creates new tracks and a hidden 
Markov model (HMM) based  visibility process that handles disappearing targets. Until here, we assumed that the existence variables $e_{tn}$ were given.In this section we present the inference modelfor the existence variable based on the stochastic birth-process.

\subsection{Birth Process}
\label{subsec:track-birth}
The principle of the person birth process is to search for consistent trajectories in the history of observations associated to clutter. Intuitively, 
two hypotheses ``\textit{the considered observation sequence is generated by a person not being tracked}'' and ``\textit{the considered 
observation sequence is generated by clutter}'' are confronted.

The model of ``\textit{the considered observation sequence is generated by a person not being tracked}'' hypothesis is based on the observations and 
dynamic models defined in \eqref{eq:motion-likelihood} and \eqref{eq:dynamical-model}. If there is a not-yet-tracked person $n$ generating the 
considered observation sequence $\{\obsval_{t-L,k_L},\ldots,\obsval_{t,k_0}\}$,\footnote{In practice we considered 
$L=2$, however, derivations are  valid for arbitrary values of L.} then the observation likelihood 
is $p(\obsval_{t-l,k_l}|\stateval_{t-l,n}) = g(\obsval_{t-l,k_l};\obsOp \stateval_{t-l,n}, \obsCov )$ and the person trajectory is governed by 
the dynamical model $p(\stateval_{t,n}|\stateval_{t-1,n}) = g(\stateval_{t,n};\trans \stateval_{t-1,n},\stateCov_n)$. Since there is no prior 
knowledge about the starting point of the track, we assume a ``flat" Gaussian distribution over $\stateval_{t-L,n}$, namely
$p_b(\stateval_{t-L,n})=g(\stateval_{t-L,n};\mathbf{m}_b,\mathbf{\Gamma}_b)$, which is approximatively equivalent to a uniform distribution over the image. 
Consequently, the joint 
observation distribution writes:
\begin{align}
\nonumber 
\tau_0 & =p (\obsval_{t,k_0},\ldots,\obsval_{t-L,k_L}) \\
\nonumber
&= \int p(\obsval_{t,k_0},\ldots,\obsval_{t-L,k_L},\stateval_{t:t-L,n})d\stateval_{t:t-L,n} \\
\label{eqn:birth-test}
&= \int \prod_{l=0}^L p(\obsval_{t,k_l}|\stateval_{t-l,n})\times \prod_{l=0}^{L-1} p(\stateval_{t-l,n}|\stateval_{t-l-1,n})\times 
p_b(\stateval_{t-2,n}) d\stateval_{t:t-L,n}, 
\end{align}
which can be seen as the marginal of a multivariate Gaussian distribution. Therefore, the joint observation distribution 
$p(\obsval_{t,k_0},\obsval_{t-1,k_1},\ldots,\obsval_{t-2,k_L})$ is also Gaussian and can be explicitly computed. 

The model of ``\textit{the considered observation sequence is generated by clutter}'' hypothesis is based on the observation model given 
in \eqref{eq:motion-likelihood}. When the considered observation sequence $\{\obsval_{t,k_0},\ldots,\obsval_{t-L,k_L}\}$ is generated by clutter, 
observations are independent and identically uniformly distributed. In this case, the joint observation likelihood is
\begin{equation}
\tau_1=p (\obsval_{t,k_0},\ldots,\obsval_{t-L,k_L}) =\prod_{l=0}^L \unif(\obsval_{t-l,k_l}).
\label{eq:tau_1}
\end{equation}

\addnote[existing-birth]{1}{Finally, our birth process is as follows: for all $\obsval_{t,k_0}$ such that $\tau_0>\tau_1$, a new person is added by 
setting $\ex_{tn}=1$}, 
$q(\stateval_{t,n};\smean_{t,n},\scov_{t,n})$ with $\smean_{t,n} = [\obsval_{t,k_0}^\top,\mathbf{0}_2^\top]^\top$, and $\scov_{tn}$ is set to the value of 
a birth covariance matrix (see \eqref{eqn:q_x}). Also, the reference appearance model for the new person is defined as 
$\histval_{t,n}=\histval_{t,k_0}$.

\subsection{Person-Visibility Process}
\label{subsec:track-visibility}
A tracked person is said to be visible at time $t$ whenever there are observations associated to that person, otherwise the 
person is considered not visible. Instead of deleting tracks, as classical for death processes, our model labels tracks without associated 
observations as \textit{sleeping}. In this way, we keep the possibility to awake such sleeping tracks when their reference appearance model highly matches an observed appearance.

We denote the $n$-th person visibility (binary) variable by $\vis_{tn}$, meaning that the person is visible at time $t$ if $\vis_{tn}=1$ and $0$ 
otherwise. We assume the existence of a transition model for the hidden visibility variable $\vis_{tn}$. More precisely, the visibility state 
temporal evolution is governed by the transition matrix, $p(\vis_{tn} =j|\vis_{t-1,n}=i)=\visprob^{\delta_{ij}} 
(1-\visprob)^{1-\delta_{ij}}$, where  $\visprob$ is the probability to remain in the same state. To enforce temporal smoothness, the probability to 
remain in the same state is taken higher than the probability to switch to another state. 

\addnote[visibility-proc]{1}{The goal now is to estimate the visibility of all the persons. For this purpose we define the visibility observations as 
$\visobs_{tn}=\ex_{tn}\sum_{i=1}^I \mult_{tn}^i$, being $0$ when no observation is associated to person $n$. In practice, we need 
to filter the visibility state variables $\vis_{tn}$ using the visibility observations $\visobs_{tn}$. In other words, we need to estimate the 
filtering distribution $p(\vis_{tn}|\visobs_{1:tn},\ex_{1:tn})$ which can be written as:
\begin{align}
p(\vis_{tn} & =\visval_{tn}|\visobs_{1:t},\ex_{1:tn})= \nonumber \\
\label{eq:visibility-filtering}
& \frac{p(\visobs_{tn}|\visval_{tn},\ex_{tn})\sum_{\visval_{t-1,n}} p(\visval_{tn}|\visval_{t-1,n}) 
p(\visval_{t-1,n}|\visobs_{1:t-1,n},\ex_{1:t-1})}{p(\visobs_{tn}|\visobs_{1:t-1,n},\ex_{1:t})},
\end{align}
where the denominator corresponds to integrating the numerator over $\visval_{tn}$. In order to fully specify the model, we define the 
visibility observation likelihood as:
\begin{equation}
p(\visobs_{tn}|\visval_{tn},\ex_{tn})=(\textrm{exp}(-\lambda \visobs_{tn}))^{\visval_{tn}}(1-\textrm{exp}(-\lambda \visobs_{tn}))^{1-\visval_{tn}} 
\end{equation}
Intuitively, when $\visobs_{tn}$ is high, the likelihood is large if $\visval_{tn}=1$ (person is visible). The opposite behavior 
is found when $\visobs_{tn}$ is small. Importantly, at each frame, because the visibility state is a binary variable, its filtering distribution can be straightforwardly computed.}
%
%
%
\section{Experiments}
\label{sec:experiments} 

\subsection{Evaluation Protocol}
\begin{figure}[t]
\begin{center}
\subfigure[CPD-2]{\label{subfig:CPD-2}\includegraphics[width=0.31\textwidth,height=3.2cm]{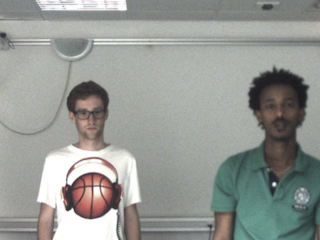} }
\subfigure[CPD-3]{\label{subfig:CPD-3}\includegraphics[width=0.31\textwidth,height=3.2cm]{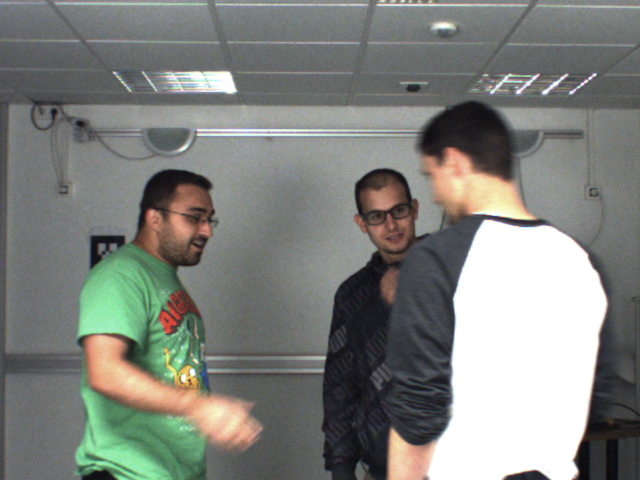}  }
\subfigure[PETS09S2L1]{\label{subfig:PETS09S2L1}\includegraphics[width=0.31\textwidth,height=3.2cm]{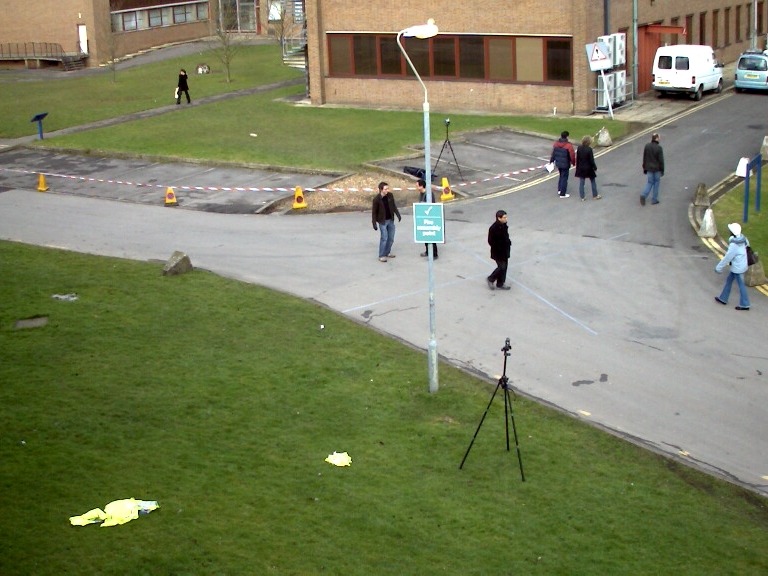}  }
\subfigure[TUD-Stadtmitte]{\label{subfig:TUD-Stadtmitte}\includegraphics[width=0.31\textwidth,height=3.2cm]{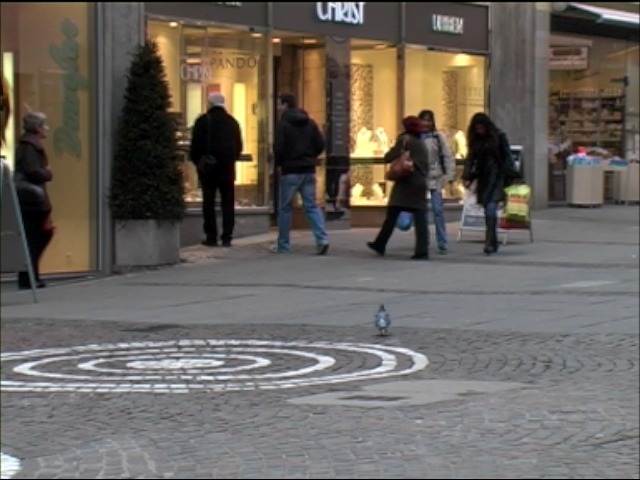}  }
\subfigure[ParkingLot]{\label{subfig:ParkingLot}\includegraphics[width=0.31\textwidth,height=3.2cm]{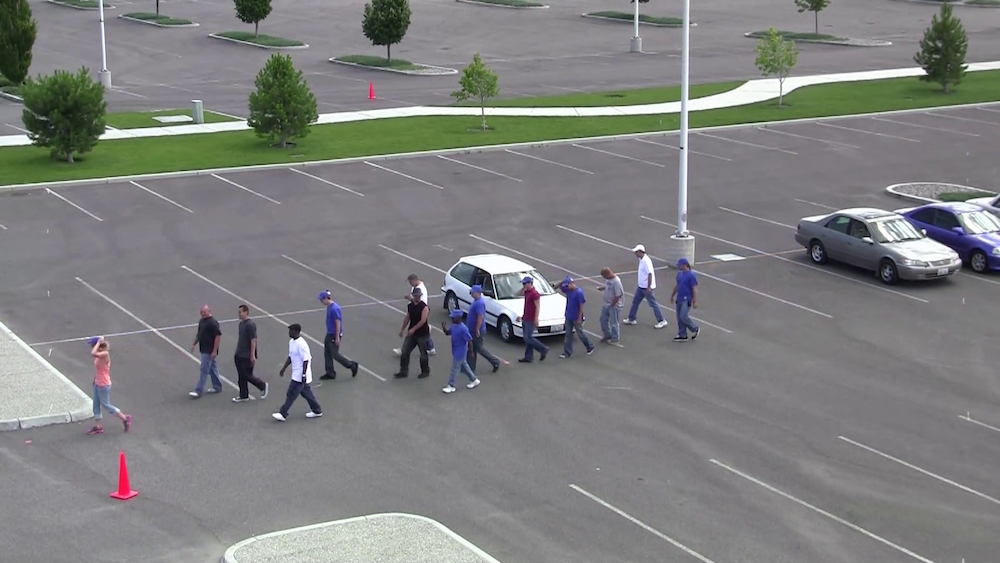}  }
\subfigure[TownCentre]{\label{subfig:TownCentre}\includegraphics[width=0.31\textwidth,height=3.2cm]{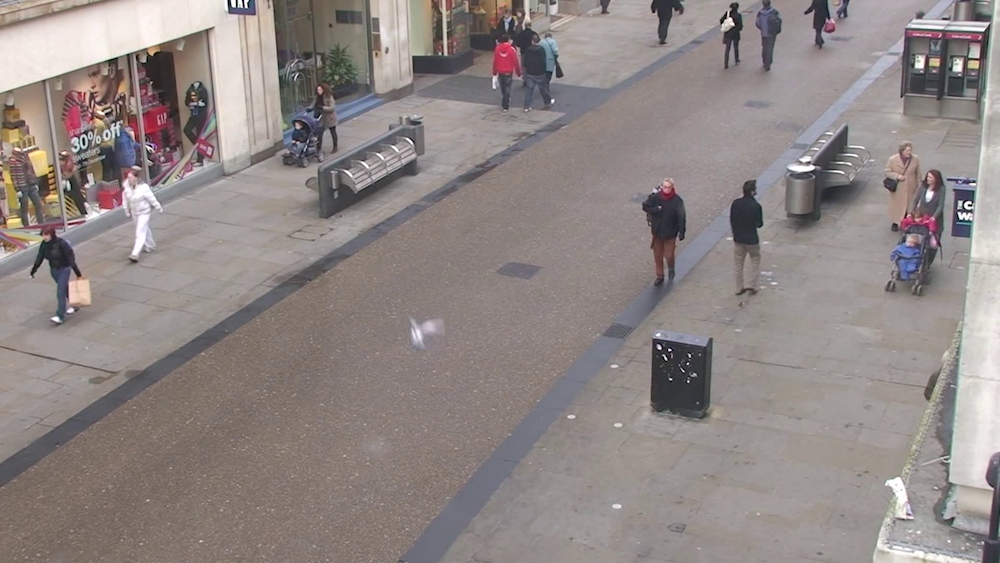}  }
\caption{Typical images extracted from the sequences used for tracking evaluation. Figures \ref{subfig:CPD-2} and \ref{subfig:CPD-3} are from the  Cocktail-Party Dataset. Figures  \ref{subfig:PETS09S2L1}, \ref{subfig:TUD-Stadtmitte}, \ref{subfig:ParkingLot}, \ref{subfig:TownCentre}
display sample images from  PETS09S2L1, TUD-Stadtmitte, ParkingLot, and TownCentre which classically used in computer vision to evaluate
multi-person tracking.}
\label{fig:eval-dataset-samples}
\end{center}
\end{figure}

We experimentally assess the performance of the proposed model using two datasets. The cocktail party dataset (CPD) is composed of two videos, CPD-2 
and CPD-3, recorded with a close-view camera (see Figure~\ref{subfig:CPD-2} and~\ref{subfig:CPD-3}). Only people's upper body is visible, and 
mutual occlusions happen often. CPD-3 records 3 persons during 853 frames and CPD-2 records 2 persons during 495 frames. 

\addnote[extra-seqs]{1}{The second dataset is constituted of four sequences classically used in computer vision to evaluate multi-person tracking 
methods~\cite{Milan-TPAMI-2014,Heili-TIP-2014}. Two  sequences were selected from the MOT Challenge Dataset \cite{MOTChallenge2015}:\footnote{\url{http://motchallenge.net/}}  TUD-Stadmitte (9 persons, 179 frames) and PETS09-S2L1 (18 persons, 795 frames). The third sequence is 
the TownCentre sequence (231 persons, 4500 frames) recorded by the Oxford Active Vision Lab. The last one is ParkingLot (14 persons, 749) recorded by the Center for Research in Computer Vision of University of Central Florida. 
TUD-Stadmitte records closely viewed full body pedestrians. PETS09-S2L1 and ParkingLot features a dozen of far-viewed full body pedestrians. TownCentre 
captures a very large number of far viewed pedestrians. This evaluation dataset is diverse and large (more than 6000 frames) enough to give a 
reliable assessment of the multi-person tracking performance measures. Figure~\ref{fig:eval-dataset-samples} shows typical views of all 
the sequences.}

Because multi-person tracking intrinsically implies track creation, deletion, target identity maintenance, and localization, evaluating multi-person 
tracking models is a non-trivial task. Many metrics have been proposed, see~\cite{RisticFUSION2010,Smith-EEMCV-2005,ClearEvaluation2006,Wen-Arxiv-2015}. In this 
paper, for the sake of completeness we use several of them split into two groups.

The first set of metrics follow the widely used CLEAR multi-person tracking evaluation metrics \cite{ClearEvaluation2006} which are commonly 
used to evaluate multi-target tracking where targets' identities are jointly estimated together with their kinematic states. On the one side the {\it 
multi-object tracking accuracy} ({\bf MOTA}) combines false positives (FP), missed targets (FN), and identity switches (ID). On the other side, the 
{\it multi-object tracking precision} ({\bf MOTP}) measures the alignment of the tracker output bounding box with the ground truth. We also provide 
tracking precision ({\bf Pr}) and recall ({\bf Rc}).

\addnote[set-metrics]{1}{The second group of metrics is specifically designed for multi-target tracking models that do not estimate the targets' 
identities, such as the PHD filter. These metrics compute set distances between the ground truth set of objects present in the scene and the set 
of objects estimated by the tracker~\cite{RisticFUSION2010}. The metrics are the {\bf Hausdorff} metric, the optimal mass transfer ({\bf 
OMAT}) metric, and the optimal sub-pattern assignment ({\bf OSPA}) metric. We will use these metrics to compare the tracking results achieved by 
our variational tracker to the results achieved by the PHD filter which does not infer identities \cite{ClarkICIF2006}.}

\addnote[complexity]{1}{The computational cost of the proposed model is mainly due the the observation extraction, namely the person detection. This process is known in computer vision to be computationally intensive. However, there are pedestrian detectors that achieve real time performances \cite{AngelovaBMVC2015}. The VEM part of the tracking model, which involves only inversion of 6 by 6 matrices, is computationally efficient and can be made real time. It converges in less than 10 steps. }

\subsection{Validation on the Cocktail Party Dataset}

\begin{figure}[t]
\begin{center}
\subfigure[]{\label{subfig:region-splitting-ub}\includegraphics[width=0.3\textwidth,height=3.7cm]{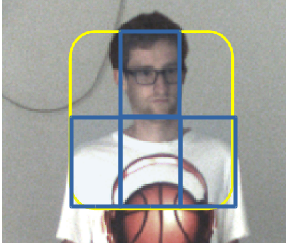} }
\subfigure[]{\label{subfig:region-splitting-b}\includegraphics[width=0.15\textwidth,height=3.7cm]{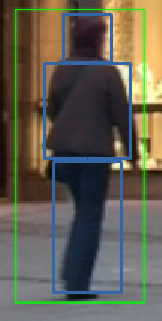}  }
\caption{Region splitting for computing the color histograms: Fig.\ref{subfig:region-splitting-ub} shows an example with upper-body detection while 
 Fig.\ref{subfig:region-splitting-b} shows an example of full body detection.}
\label{fig:histogram-region-splitting}
\end{center}
\end{figure}

In the cocktail party dataset our model exploits upper body detections obtained using~\cite{FelzenszwalbTPAMI2010} and face detections obtained 
using~\cite{ZhuRamananCVPR2012}. Therefore, we have two types of observations, upper body \textsc{u} and face \textsc{f}. The hidden 
state corresponds to the position and velocity of the upper body. The observation operator $\obsOp^{\textsc{u}}$ (see section 
\ref{subsubsec:observation-likelihood}) for the upper body observations simply removes the velocity components of the hidden state. The observation 
operator $\obsOp^{\textsc{f}}$ for the face observations combines a projection removing the velocity components and an affine mapping (scaling and translation) transforming  face localization bounding boxes into the the upper body localization bounding boxes. The appearance observations are concatenations of joint hue-saturation color histograms of the torso 
split into three different regions, plus the head region as shown in Fig.\ref{subfig:region-splitting-ub}.

Tables~\ref{tab:performances-cocktail-party} and~\ref{tab:set-metric-performances-cpd} show the performance of the model over the two sequences of 
the cocktail party dataset. While in Table~\ref{tab:performances-cocktail-party} we evaluate the performance of our model under the first set of 
metrics, in Table~\ref{tab:set-metric-performances-cpd}, we compare the performance of our model to the one of the GMM PHD filter using the set-based 
metrics. Regarding the detectors, we evaluate the performance when using (i)~upper body detectors, (ii)~face detectors or (iii)~both. For each of 
these three choices, we also compare when adding color histogram descriptors or when not using them. From now on, \textsc{u} and \textsc{f} denote 
the use of upper-body detectors and face detectors respectively, while \textsc{c} denotes the use of color histograms.

\begin{table}[t]
\begin{center}
\begin{tabular}{llcccc}
\toprule
Sequence & Features & {\bf Rc}   & {\bf Pr}    & {\bf MOTA} & {\bf MOTP}  \\
\midrule
\multirow{3}{*}{CPD-2} &   \textsc{u}/\textsc{uc}    & 53.3/70.7 & 94.9/99.4 & 46.6/64.3 & 80.8/85.8\\ 
                       &  \textsc{f}/\textsc{fc}     & 89.8/90.1 & 94.6/94.6 & 75.7/76.0 & 76.6/76.7\\
                       &  \textsc{fu}/\textsc{fuc}   & 93.1/95.2 & 95.3/96.2 & 88.3/80.0 & 76.5/82.9 \\
\midrule
\multirow{3}{*}{CPD-3} &   \textsc{u}/\textsc{uc}   & 93.6/93.6 & 94.4/99.6 & 91.6/91.8 & 85.0/86.8\\
			           &   \textsc{f}/\textsc{fc}   & 62.5/62.8 & 97.6/98.4 & 58.9/59.7 & 68.5/68.4\\
			          &   \textsc{fu}/\textsc{fuc}  & 91.0/92.6 & 99.4/99.7 &  88.3/90.1 & 76.5/82.9\\
\bottomrule
\end{tabular}
\end{center}
\caption{Evaluation of the proposed multi-person tracking method with different features on the two sequences of the 
cocktail party dataset. All measures are in \%.}
\label{tab:performances-cocktail-party}
\end{table}

Results in Table~\ref{tab:performances-cocktail-party} show that for the sequence CPD-2, while {\bf Pr} and {\bf MOTP} are higher when using 
upper-body detections \textsc{u/uc}, {\bf Rc} and {\bf MOTA} are higher when using face detections \textsc{f/fc}. One may think that the 
representation power of both detections may be complementary to each other. This is evidenced in the third row of 
Table~\ref{tab:performances-cocktail-party}, where both detectors are used and the performances are higher than in the first two rows, except for 
{\bf Pr} and {\bf MOTP} when using color. Regarding CPD-3, we clearly notice that the use of upper-body detections is much more advantageous than 
using the face detector. Importantly, even if the performance reported by the combination of the two detectors does not significantly outperform the 
ones reported when using only the upper-body detectors, it exhibits significant gains when compared to using only face detectors. The use of color 
seems to be advantageous in most of the cases, independently of the sequence and the detections used. Summarizing, while the performance of the 
method using only face detections or upper-body detections seems to be sequence-dependent, there is a clear advantage of using the feature 
combinations. Indeed, the combination seems to perform comparably to the best of the two detectors and much better to the worst. Therefore, the use of 
the combined detection appears to be the safest choice in the absence of any other information and therefore justifies developing a model able to 
handle observations coming from multiple detectors.

\begin{table*}[ht]
\begin{center}
\begin{tabular}{llccc}
\toprule
Sequence & Method-Features & {\bf Hausdorff}  & {\bf OMAT}    & {\bf OSPA} \\
\midrule
\multirow{6}{2cm}{CPD-2} &   VEM-\textsc{u/uc}     & 239.4/239.2   & 326.5/343.1   & 247.8/244.5  \\
                         &   PHD-\textsc{u}        & 276.6 & 435.3 & 567     \\
                         &   VEM-\textsc{f/fc}      & 116.3/115.5   & 96.3/96.1    & 110.9/108.0 \\
                         &   PHD-\textsc{f}         & 124 & 102 & 185.8  \\
                         &   VEM-\textsc{fu/fuc}   & 98.0/97.7   & 80.3/7    & 92.7/90.6  \\ 
                         &   PHD-\textsc{fu}       & 95  & 80 & 168    \\
\midrule
\multirow{6}{2cm}{CPD-3} &   VEM-\textsc{u/uc}     & 56.0/56.2  &  44.4/44.2  &  54.7/54.1 \\
                         &   PHD-\textsc{u}        & 162.2   & 244.6   &  382.6  \\
                         &   VEM-\textsc{f/fc}     & 184.2/185.5  & 200.8/201.3436 & 203.3/205.0 \\
                         &   PHD-\textsc{f}        & 208     & 239.5   & 445.2   \\
			            &   VEM-\textsc{fu/fuc}   &  66.3/67.4  & 52.7/52.8  & 68.5/68.0 \\
			            &   PHD-\textsc{fu}       & 49     & 54.4   & 181   \\
\bottomrule
\end{tabular}
\end{center}
\caption{Set metric based multi-person tracking performance measures of the proposed VEM and of the GMM PHD filter  \cite{ClarkICIF2006} on the the cocktail party dataset.}
\label{tab:set-metric-performances-cpd}
\end{table*}

Table~\ref{tab:set-metric-performances-cpd} reports a comparison of the proposed VEM model with the PHD filter for different features under the 
set metrics over the two sequences of the cocktail party dataset. We first observe that the behavior described from the results of 
Table~\ref{tab:performances-cocktail-party} is also observed here, for a different group of measures and also for the PHD filter. Absolutely, while 
the use of the face or of the upper-body detections may be slightly more advantageous than the combination of detectors, this is sequence- and 
measure-dependent. However, the gain of the combination over the less reliable detector is very large, thus justifying the multiple-detector strategy 
when the applicative scenario allows for it and no other information about the sequence is available. The second observations is that the proposed 
VEM outperforms the PHD filter almost everywhere (i.e. except for CDP-3 with \textsc{fu/fuc} under the Hausdorff measure). This systematic trend 
demonstrates the potential of the proposed  method from an experimental point of view. One possible explanation maybe that the variational tracker 
exploits additional information as it jointly estimates the target kinematic states together with their identities.
\begin{figure*}[t]
\centering
\subfigure[CPD-2]{\includegraphics[width=0.45\textwidth]{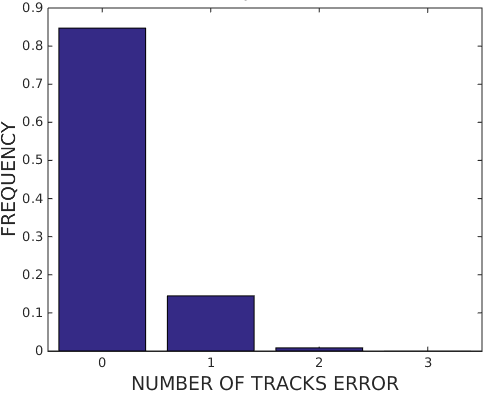} }
\subfigure[CPD-3]{\includegraphics[width=0.45\textwidth]{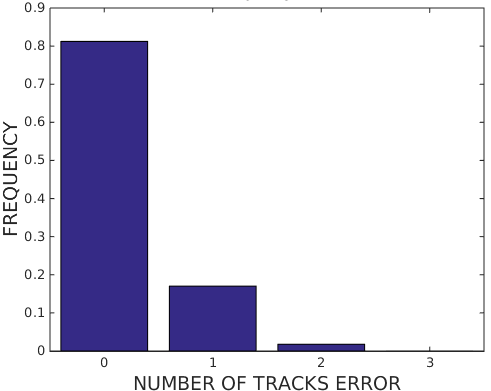} }
\caption{Histogram of absolute errors about the estimation of the number of people present in the visual scene over the Cocktail Party Dataset.}
\label{fig:nb_tracks_errors_CPD}
\end{figure*}

Figure \ref{fig:nb_tracks_errors_CPD} gives the histograms of the number of persons estimation absolute errors made by the variational tracking model.
These results shows that for over the Cocktail Party Dataset, the number of people present in the visual scene for in a given time frame are in general
correctly estimated. This shows that birth and the visibility processes play their role in creating tracks when new people enter the scene, and when they 
are occluded or leave the scene. More than $80\%$ of the time, the correct number of people is correctly estimated. It has to be noticed that errors 
are slightly higher for the sequence involving three person than for the sequence involving two persons.

To give a qualitative flavor to the tracking performance, Figure \ref{fig:tracking-PETS09S2L1} gives sample results achieved by the proposed model 
(VEM-\textsc{fuc}) on CPD-3. These images show that the model is able to correctly initialize new tracks, identify occluded people as no longer 
visible, and recover their identities after occlusion. Tracking results are provided as supplementary material.

\begin{figure*}[t]
\centering
\includegraphics[width=0.240\textwidth]{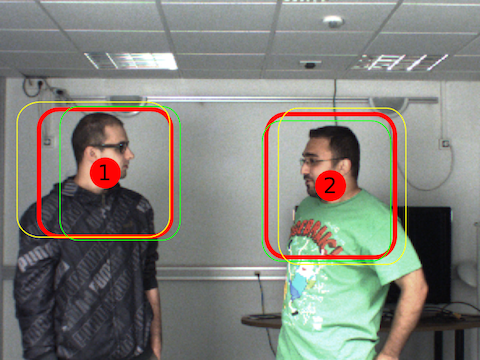}
\includegraphics[width=0.240\textwidth]{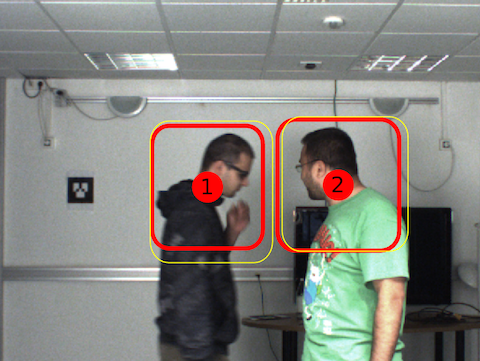}
\includegraphics[width=0.240\textwidth]{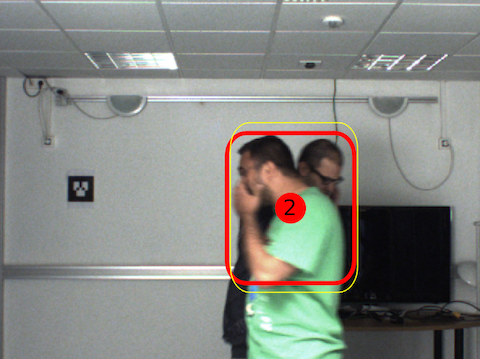}
\includegraphics[width=0.240\textwidth]{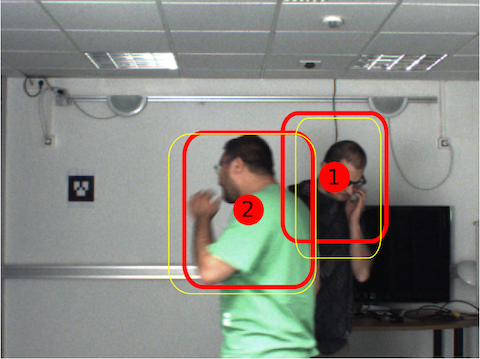}

\includegraphics[width=0.240\textwidth]{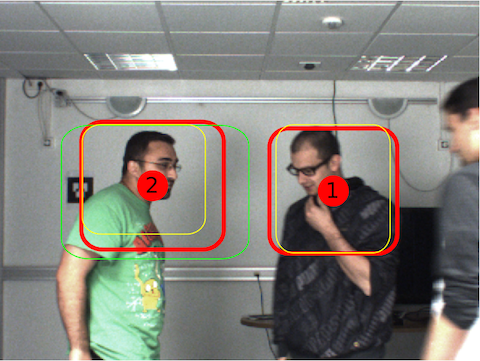}
\includegraphics[width=0.240\textwidth]{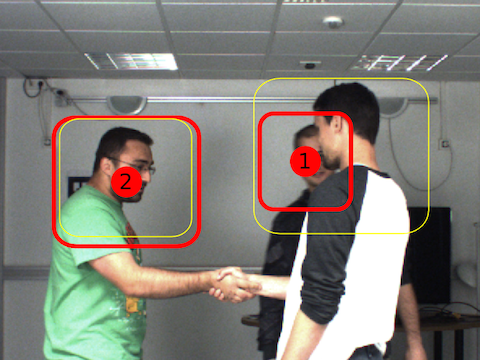}
\includegraphics[width=0.240\textwidth]{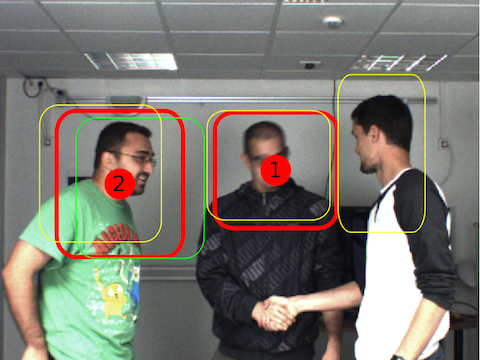}
\includegraphics[width=0.240\textwidth]{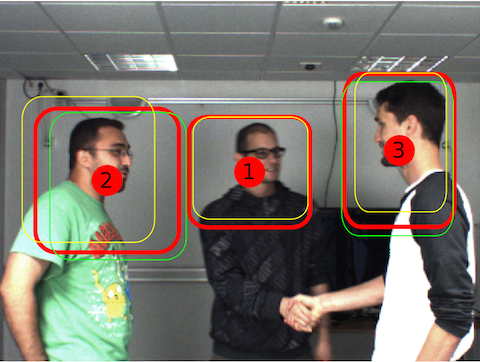}
\caption{Sample tracking results on CPD-3. The green bounding boxes represent the face detections and the yellow bounding boxes represent the upper 
body detections. Importantly, the red bounding boxes display the tracking results.}
\label{fig:tracking-CP-seq1}
\end{figure*}

\begin{figure*}[t]
\centering
\includegraphics[width=0.8\textwidth]{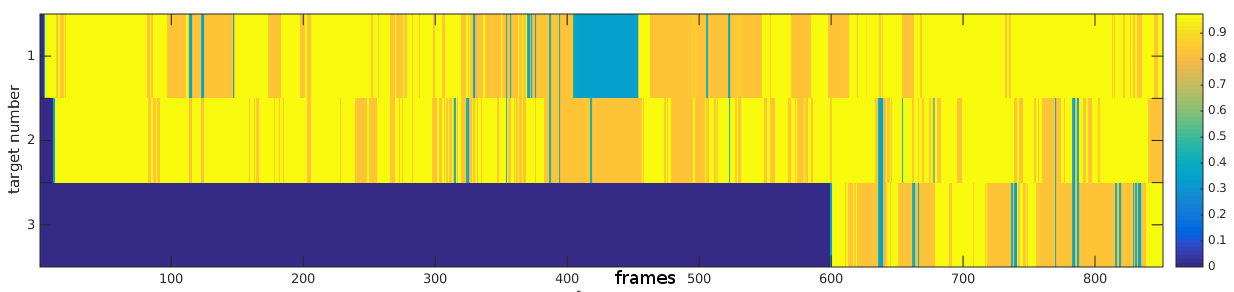}
\caption{Estimated visibility probabilities for tracked persons in sequence CPD-3. Every row displays the corresponding targets visibility 
probabilities for every time frame. Yellow color represents very high probability (close to 1), and blue color represents very low probabilities.}
\label{fig:tracking-CP-seq1-visibility}
\end{figure*}

Figure \ref{fig:tracking-CP-seq1-visibility} gives the estimated targets visibility probabilities (see Section \ref{subsec:track-visibility}) for 
sequence CPD-3 with sample tracking images given in Figure \ref{fig:tracking-CP-seq1}. The person visibility show that tracking for person 1 and 2 
starts at the beginning of the sequence, and person 3 arrives at frame 600. Also, person 1 is occluded between frames 400 and 450 (see fourth image 
in the first row, and first image in the second row of Figure \ref{fig:tracking-CP-seq1}). 

\subsection{Evaluation on classical computer vision video sequences}

In this tracking situation, we model a single person's kinematic state as the full body bounding box and its velocity. In this case, the observation 
operator $\obsOp$ simply removes the velocity information, keeping only the bounding box' position and size. The appearance observations are the 
concatenation of the joint HS histograms of the head, torso and legs areas (see Figure \ref{subfig:region-splitting-b}).

\begin{table}[ht]
\centering
\begin{tabular}{llccc}
\toprule
Sequence                             & Method-Features & {\bf Hausdorff}   & {\bf OMAT}    & {\bf OSPA} \\
\midrule
\multirow{2}{3.5cm}{TUD-Stadtmitte } &  VEM-\textsc{b/bc}      & 150.4/125.9 &   197.5/184.9    &  483.2/482.4\\
                                     &  PHD-\textsc{b}         & 184.7       &   119            &  676 \\
\midrule
\multirow{2}{3.5cm}{PETS09S2L1 }     &   VEM-\textsc{b/bc}     & 52.1/50.9   &  72.6/40.8       &  117.0/110.1\\
                                     &   PHD-\textsc{b}        & 70          &  44              &  163 \\
\midrule
\multirow{2}{3.5cm}{TownCentre}      &   VEM-\textsc{b/bc}     &  420./391.2 &  205.4/177.5     &350.0 /335.2      \\
                                     &   PHD-\textsc{b}        &  430.5      &  173.8           &  364.9   \\
\midrule
\multirow{2}{3.5cm}{ParkingLot}      &   VEM-\textsc{b/bc}     & 95.0/90.5       &  87.9/83.9           & 210.8/203.4 \\
                                     &   PHD-\textsc{b}        & 169         &  94.0       & 415  \\
\bottomrule
\end{tabular}
\caption{Set metric based multi-person tracking Performance measures on the sequences the four sequences PETS09S2L1, TownCentre, ParkingLot,and TUD-Stadtmitte.}
\label{tab:set-metric-performances-MOT-Challenge}
\end{table}

We evaluate our model using only body localization observations (\textsc{b}) and jointly using body localization and color appearance observations 
(\textsc{bc}). Table~\ref{tab:set-metric-performances-MOT-Challenge} compare the proposed variational model to the PHD filter using set based 
distance performance metrics. As for the cocktail party dataset, in general, these results show that the variational tracker outperforms the PHD 
filter.

In addition, we also compare the proposed model to two tracking models, proposed by Milan {\it et al} in \cite{Milan-TPAMI-2014} and by Bae and Yoon 
in \cite{BaeYoonCVPR2014}. Importantly, the direct comparison of our model to these two state-of-the-art methods must be done with care. Indeed, 
while the proposed VEM uses only causal (past) information, these two methods use both past and future detections. In other words, while ours is 
a \textit{filtering},  \cite{Milan-TPAMI-2014,BaeYoonCVPR2014} are \textit{smoothing} methods. Therefore, we expect these two models to outperform 
the proposed one. However, the main prominent advantage of filtering methods over smoothing methods, and therefore of the proposed VEM over these two 
methods, is that while smoothing methods are inherently unsuitable for on-line processing, filtering methods are naturally appropriate for on-line 
task, since they only use causal information.

\begin{table}[t]
\hspace{-0.6cm}
\centering
\begin{tabular}{llcccc}
\toprule
Sequence                                    & Method & Rc    & Pr    &  MOTA & MOTP \\
\midrule
\multirow{3}{3cm}{TUD-Stadmitte}            &  VEM-\textsc{b/bc}      & 72.2/70.9   & 81.7/82.5  &  54.8/53.5 & 65.4/65.1 \\
                                            & \cite{Milan-TPAMI-2014} & 84.7        & 86.7       & 71.5       & 65.5 \\
\midrule
\multirow{3}{3cm}{PETS09-S2L1}              &  VEM-\textsc{b/bc}      & 90.1/90.2   & 86.2/87.6  & 74.9/76.7  & 71.8/71.8 \\
                                            & \cite{Milan-TPAMI-2014} & 92.4        & 98.4       & 90.6       & 80.2\\
                                            & \cite{BaeYoonCVPR2014}  &  -          & -          & 83         & 69.5 \\
\midrule
\multirow{1}{3cm}{TownCentre}               & VEM-\textsc{b/bc}       &   88.1/90.1  & 71.5/72.7 & 72.7/70.9  & 74.9/76.1  \\
    
\midrule
\multirow{1}{3cm}{ParkingLot}               & VEM-\textsc{b/bc}       &   80.3/78.3   & 85.2/87.5   & 73.1/74 & 70.8/71.7  \\ 
\bottomrule
\end{tabular}
\caption{Performance measures on the sequences of the second dataset. Comparison with \cite{Milan-TPAMI-2014,BaeYoonCVPR2014} must be done with care 
since both are smoothing methods and therefore use more information than the proposed VEM.}
\label{tab:performances-motchallenge}
\end{table}

Table \ref{tab:performances-motchallenge} reports the performance of these methods on four sequences classically used in computer vision to evaluate 
multi-target trackers. In this table, results over TUD-Stadmitte show similar performances for our model using or not appearance information. 
Therefore, color information is not very informative in this sequence. In PETS09-S2-L1, our model using color achieves better MOTA measure, precision, 
and recall, showing the benefit of integrating color into the model. As expected, Milan {\it et al} and Bae and Yoon, outperform the proposed model. 
However, the non-causal nature of their method makes them unsuitable for on-line tracking tasks, where the observations must be processed when 
received, and not before. 
\begin{figure*}[t]
\centering
\subfigure[PETS09-S2L1]{\includegraphics[width=0.45\textwidth]{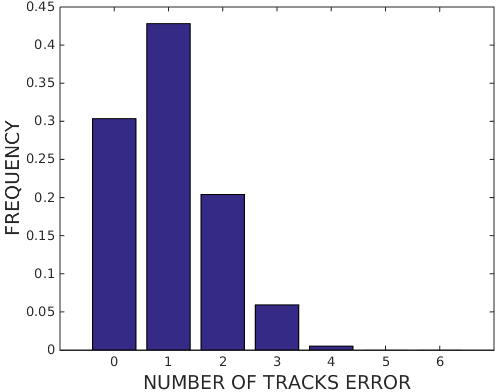} }
\subfigure[TownCentre]{\includegraphics[width=0.45\textwidth]{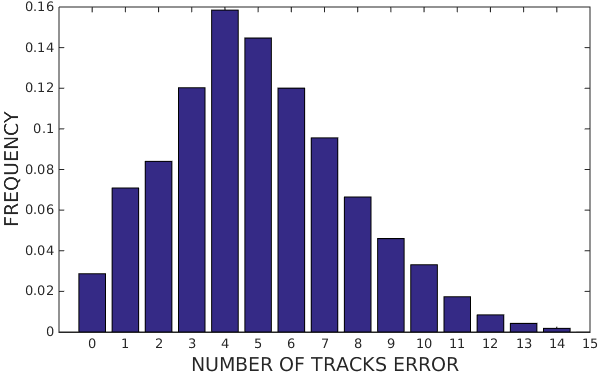} }
\subfigure[ParkingLot]{\includegraphics[width=0.45\textwidth]{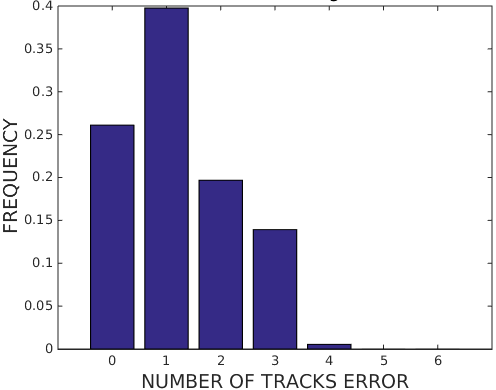} }
\subfigure[TUD-Stadtmitte]{\includegraphics[width=0.45\textwidth]{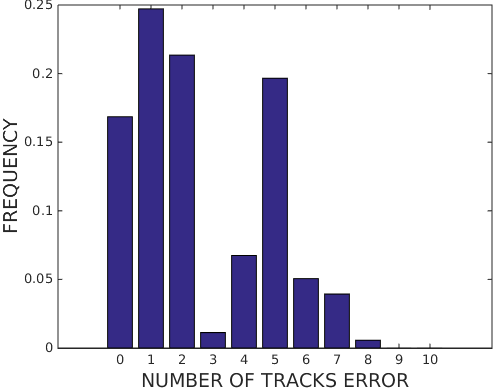} }
\caption{Histogram of errors about the estimation of the number of people present in the visual scene over ParkingLot, TownCentre, PETS09-S2L1, TUD-Stadtmitte.}
\label{fig:nb_tracks_errors_Classical}
\end{figure*}

Figure \ref{fig:nb_tracks_errors_Classical} gives the histograms of the errors about the number of people present in the visual scene for the four sequences
 ParkingLot, TownCentre, PETS09-S2L1, TUD-Stadtmitte. These results show that, the four sequences are more challenging than the Cocktail Party Dataset 
 (see figure \ref{fig:nb_tracks_errors_CPD}). Among the four video sequences, TUD-Stadtmitte is the one where variational tracking model is making the estimated number of people is the less consistent. This can be explained by the quality of the observations (detections) over this sequence. For the PETS, and the ParkingLot dataset which involve about 15 persons, about 70\% of the time the proposed tracking model is estimating the number of people in the scene with an error below 2 persons. For the TownCentre sequence which involves 231 persons over 4500 frames, over 70\% of the time, the error made by the variational tracker is below 7 persons. This shows that, even in challenging situations involving occlusions due to crowd, the birth and the visibility process play their role.

\begin{figure*}[p!]
\centering
\includegraphics[width=0.40\textwidth]{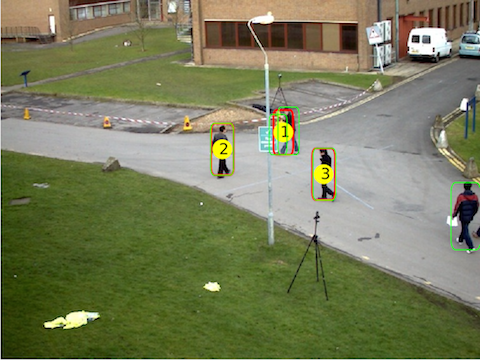}\hspace{0.4mm}%
\includegraphics[width=0.40\textwidth]{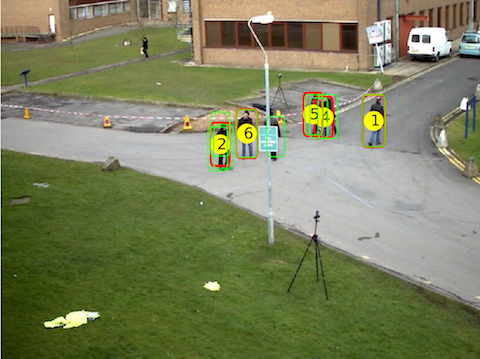}\\
\includegraphics[width=0.40\textwidth]{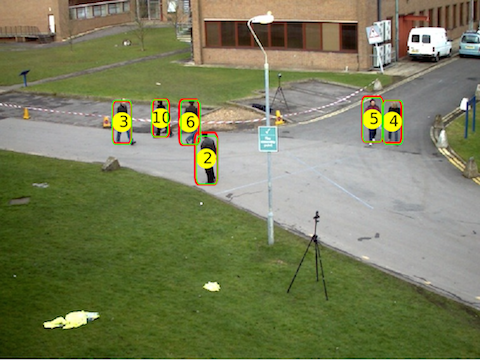}\hspace{0.4mm}%
\includegraphics[width=0.40\textwidth]{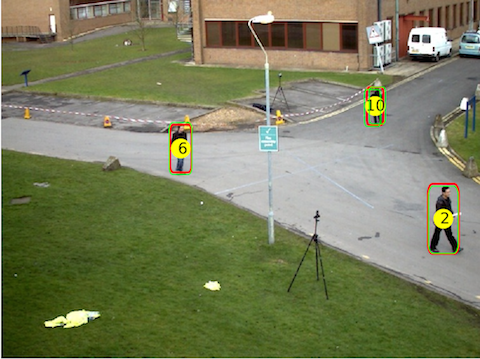}\\
\includegraphics[width=0.40\textwidth]{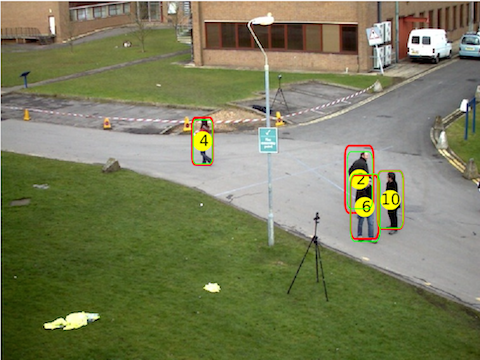}\hspace{0.4mm}%
\includegraphics[width=0.40\textwidth]{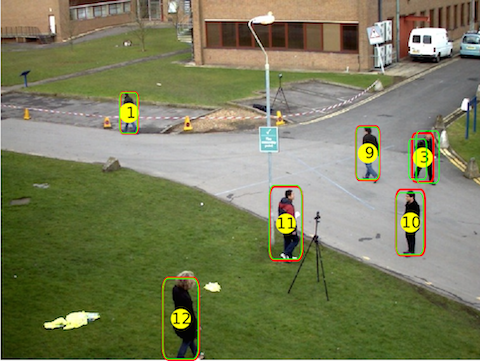}\\
\includegraphics[width=0.40\textwidth]{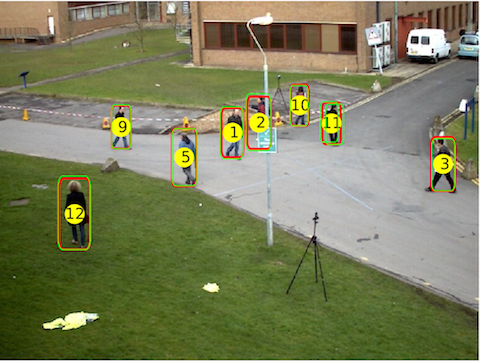}\hspace{0.4mm}%
\includegraphics[width=0.40\textwidth]{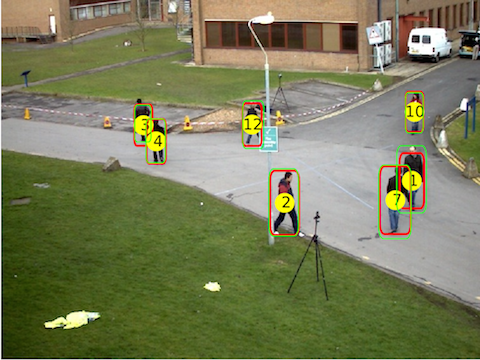}
\caption{Tracking results on PETS09-S2L1. Green boxes represent observations  and red bounding 
boxes represent tracking outputs associated with person identities. Green and red bounding boxes may overlap.}
\label{fig:tracking-PETS09S2L1}
\end{figure*}

Figure \ref{fig:tracking-PETS09S2L1} presents sample results for the PET09-S2L1 sequence. In addition, videos presenting the results on the second 
dataset are provided as supplementary material. These results show temporally consistent tracks. Occasionally, person identity switches 
may occur when two people cross. Remarkably, because the proposed tracking model is allowed to reuse the identity of persons visible in the past, 
people re-entering the scene after having left, will be recognized the the previously used track will be awaken.

%
%
%
\section{Conclusions}
\label{sec:conclusions}
We presented an on-line variational Bayesian model to track a time-varying number of persons from cluttered multiple visual observations. Up 
to our knowledge, this is the first variational Bayesian model for tracking multiple persons, or more generally, multiple targets. We proposed 
birth and visibility processes to handle persons that are entering and leaving the visual field. The proposed model is evaluated with two datasets showing competitive 
results with respect to state of the art multi-person tracking models. Remarkably, even if in the conducted experiments we model the visual 
appearance with color histograms, our framework is versatile enough to accommodate other visual cues such as texture, feature descriptors or motion 
cues.  

 \addnote[birth-reference]{2}{
In the future we plan to consider the integration of more sophisticated birth processes than the one considered in this paper, e.g. \cite{Streit-SDF-2013}.} We also plan to extend the visual tracker to incorporate auditory cues. For this purpose, we plan to jointly track the kinematic states
and the speaking status (active/passive) of each tracked person. The framework proposed in this paper allows to exploit audio features, e.g.  voice activity detection and audio-source 
localization as observations. When using audio information, robust voice descriptors (the acoustic equivalent of visual appearance) and their blending with the 
tracking model will be investigated. We also plan to extend the proposed formalism to a moving camera such that its kinematic state is  
tracked as well. This case is of particular interest in applications such as pedestrian tracking for self-driving cars or for human-robot interaction.
\appendix

\section{Derivation of the Variational Formulation}

\subsection{Filtering Distribution Approximation}
\label{subsec:filtering-distribution-app}
The goal of this section is to derive an approximation of the hidden-state filtering distribution $p(\latentset_t,\state_t|\obsjset_{1:t}, \exvect_{1:t})$, given the variational approximating distribution $q(\latentset_{t-1},\state_{t-1})$ at $t-1$.
Using Bayes rule, the filtering distribution can be written as
\begin{equation} \label{eq:filtering-distribution}
 p(\latentset_t,\state_t|\obsjset_{1:t}, \exvect_{1:t}) =\frac{p(\obsjset_t|\latentset_t,\state_t,\exvect_t)p(\latentset_t,\state_t|\obsjset_{1:t-1},\exvect_{1:t})}
 {p(\obsjset_t|\obsjset_{1:t-1}, \exvect_{1:t})}.
\end{equation}
It is composed of three terms, the likelihood $p(\obsjset_t|\latentset_t,\state_t,\exvect_t)$, the predictive distribution $p(\latentset_t,\state_t|\obsjset_{1:t-1},\exvect_{1:t})$, and the normalization factor $p(\obsjset_t|\obsjset_{1:t-1}, \exvect_{1:t})$ which is independent of the hidden variables.
The likelihood can be expanded as:
\begin{eqnarray} \label{eqn:expanded-likelihood}
 p(\obsjset_t|\latentset_t,\state_t,\exvect_t)  &=&\prod_{i=1}^I \prod_{k\leq K_t^i} \prod_{n=0}^N p(\obsjval_{tk}|\latent_{tk}=n,\state_t,\exvect_t) ^{\delta_n(\latent_{tk}^i)}
 \end{eqnarray}
where $\delta_n$ is the Dirac delta function, and $p(\obsjval_{tk}|\latent_{tk}=n,\state_t,\exvect_t)$ is the individual observation likelihood defined in  \eqref{eq:motion-likelihood} and \eqref{eqn:color-likelihood}. 

The predictive distribution factorizes as 
$$p(\latentset_t,\state_t|\obsjset_{1:t-1}, \exvect_{1:t})= p(\latentset_t|\exvect_t) p(\state_t|\obsjset_{1:t-1}, \exvect_{1:t}).$$
Exploiting its multinomial nature, the assignment variable distribution $p(\latentset_t|\exvect_t)$ can be fully expanded as:
\begin{equation} \label{eqn:predictive-latent}
p(\latentset_t|\exvect_t) = \prod_{i=1}^I \prod_{k\leq K_t^i} \prod_{n=0}^N p(\latent_{tk}^i=n|\exvect_t)^{\delta_n(\latent_{tk}^i)}.
\end{equation}

Using  the motion state dynamics definition $p(\stateval_{tn}|\stateval_{t-1n},\exvect_{tn})$ the previous time motion state filtering  distribution variational approximation $q(\stateval_{t-1n}|\exvect_{t-1}) = p(\stateval_{t-1n}|\obsjset_{1:t-1},\exvect_{1:t-1})$ 
 defined in (\ref{eqn:q_x}), motion state predictive distribution $p(\state_t=\stateval_t|\obsjset_{1:t-1}, \exvect_{1:t})$ can approximated by 
 \begin{align}
p & (\state_t=\stateval_t|\obsjset_{1:t-1}, \exvect_{1:t}) \nonumber \\
 &= \int p(\stateval_t|\stateval_{t-1},\exvect_{t})p(\stateval_{t-1}|\obsjset_{1:t-1},\exvect_{1:t-1})d\stateval_{t-1}  \nonumber \\
 &= \int \left( \prod_{n=1}^N  p(\stateval_{tn}|\stateval_{t-1n},\exvect_{tn}) \right) p(\stateval_{t-1}|\obsjset_{1:t-1},\exvect_{1:t-1})d\stateval_{t-1}  \nonumber \\
 & \approx \int \prod_{n=1}^N  p(\stateval_{tn}|\stateval_{t-1n},\exvect_{tn})q(\stateval_{t-1n}|\exvect_{t-1n})d\stateval_{t-1,1}...d\stateval_{t-1,n}  \nonumber \\
 \label{eqn:predictive-motion}
&\approx \prod_{n=1}^N \unif(\stateval_{tn}) ^ {1-\ex_{tn}} g(\stateval_{tn},\trans \smean_{t-1,n},\trans \scov_{tn} \trans^\top + \stateCov_n)^{\ex_{tn}}
 \end{align}
where during the derivation, the filtering distribution of the kinematic state at time $t-1$ is replaced by its variational approximation $p(\stateval_{t-1}|\obsjset_{1:t-1},\exvect_{1:t-1}) =  \prod_{n=1}^N q(\stateval_{t-1n}|\exvect_{t-1n})$.

Equations \eqref{eqn:expanded-likelihood}, \eqref{eqn:predictive-latent}, and \eqref{eqn:predictive-motion} define the numerator of the tracking 
filtering distribution~\eqref{eq:filtering-distribution}. The logarithm of this filtering distribution is used by the proposed variational EM 
algorithm.

\label{subsec:E-steps-app}
\subsection{Derivation of the E-Z-Step}
\label{subsubsec:E-steps-assignment-app}
The E-Z-step corresponds to the estimation of $\q(\latent_{tk}^i|\exvect_t)$ given by \eqref{eqn:variational-expectations-assign}
which, from the log of the filtering distribution, can be written as:
\begin{align}
 \log  & \q (\latent_{tk}^i|\exvect_t) = \nonumber \\
    &  \sum_{n=0}^N \delta_n(\latent_{tk}^i) \mathbf{E}_{q(\state_{tn}|\exvect_t)}[ \log \left( p(\obsval_{tk}^i,\histval_{tk}^i|\latent_{tk}^i=n,\state_t,\exvect_t) p(\latent_{tk}^i=n|\exvect_t) \right)] + C,
\end{align}
where $C$ gathers terms that are constant with respect to the variable of interest, $\latent_{tk}^i$ in this case. 
By substituting $p(\obsval_{tk}^i,\histval_{tk}^i|\latent_{tk}^i=n,\state_t,\exvect_t)$, and $p(\latent_{tk}^i=n|\exvect_t)$ with their expressions \eqref{eq:motion-likelihood}, \eqref{eqn:color-likelihood}, and \eqref{eqn:assignment-distribution}, by introducing the notations
\begin{align*}
\epsilon_{tk0}^i &=  \unif(\obsval_{tk}^i)\unif(\histval_{tk}^i) \\ 
\epsilon_{tkn}^i &= g(\obsval_{tk}^i,\obsOp \smean_{tn},\obsCov^i) \exp(-\frac{1}{2} \trace({\obsOp^\top \obsCov^i}^{-1} \obsOp \scov_{tn})) b(\histval_{tk}^i,\histval_n)
\end{align*}
and after some algebraic derivations, the distribution of interest can be written as the following multinomial distribution 
\begin{equation}
\q(\latent_{tk}^i=n|\exvect_t) = \multq_{tkn}^i=\frac{ \ex_{tn}  \epsilon_{tkn}^i }{\sum_{m=0}^N \ex_{tm} \epsilon_{tkm}^i}
\end{equation}

\subsection{Derivation of the E-X-Step}
\label{subsubsec:E-step-motion-app}
 The E-step for the motion state variables consists in the estimation of $\q(\state_{tn}|\ex_{tn})$ using relation $\log \q(\state_{tn}|\ex_{tn}) = \mathbf{E}_{q(\latent_t,\state_t/\state_{tn}|\exvect_t)}[\log p(\latentset_t,\state_t|\obsjset_{1:t},\exvect_{1:t})]$ which can be expanded as 
\begin{align*}
 \log \q(\state_{tn}|\exvect_t) &=  \sum_{i=1}^I \sum_{k=0}^{K_t^i} \mathbf{E}_{q(\latent_{tk}^i|\exvect_t)}[\delta_n(\latent_{tk}^i)]\log g(\obsval_{tk}^i;\obsOp\state_{tn},\obsCov^i)^{\ex_{tn}} \\ 
 &+ \log (\unif(\state_{tn})^{1-\ex_{tn}} g(\state_{tn};\trans \smean_{t-1n},\trans \scov_{tn} \trans^\top + \stateCov_n)^{\ex_{tn}}) + C,
 \end{align*}
where, as above, $C$ gathers constant terms.
After some algebraic derivation one obtains  $\q(\state_{tn}|\ex_{tn}) = \unif(\state_{tn})^{1-\ex_{tn}} g(\state_{tn};\smean_{tn},\scov_{tn} )^{\ex_{tn}} $ where the mean and covariance of the Gaussian distribution are given by \eqref{eqn:q-motion-parameters-mean} and by \eqref{eqn:q-motion-parameters-cov}.

\bibliographystyle{elsarticle-num}

\end{document}